%% file: main.tex
\documentclass[10pt,twocolumn,letterpaper]{article}

\usepackage{iccv}
\usepackage{times}
\usepackage{epsfig}
\usepackage{graphicx}
\usepackage{amsmath}
\usepackage{amssymb}

\usepackage{booktabs,tabulary,pifont,multirow,subfig,xspace,xcolor,makecell}
\usepackage{amsthm,fixmath,mathtools,nicefrac,mmstyle}
\usepackage[titletoc]{appendix}
\usepackage[misc]{ifsym}

\newcommand{\cmark}{\ding{51}}
\newcommand{\xmark}{\ding{55}}
\newcommand{\gxmark}{\textcolor{black!20}{\ding{55}}}

\newcommand{\apr}{AP$_\textrm{r}$\xspace}
\newcommand{\apc}{AP$_\textrm{c}$\xspace}
\newcommand{\apf}{AP$_\textrm{f}$\xspace}
\newcommand{\appool}{AP$^{\text{Pool}}$\xspace}

\newcommand{\oldap}{AP$^\text{Old}$\xspace}
\newcommand{\fixedap}{AP$^\text{Fixed}$\xspace}

\definecolor{demphcolor}{RGB}{90,90,90}
\newcommand{\demph}[1]{\textcolor{demphcolor}{#1}}
\newcommand{\dt}[1]{\fontsize{6pt}{0.1em}\selectfont \demph{(#1)}}
\newcommand{\tablestyle}[2]{\setlength{\tabcolsep}{#1}\renewcommand{\arraystretch}{#2}\centering\footnotesize}
\newcolumntype{x}[1]{>{\centering\arraybackslash}p{#1pt}}
\newlength\savewidth
\newcommand{\td}[1]{\scriptsize\rlap{ \dt{#1}}}
\newcommand{\smallsec}[1]{\vspace{0.5ex}\noindent{\textbf{#1}}}

\DeclareUnicodeCharacter{2212}{-}

\usepackage[pagebackref=true,breaklinks=true,letterpaper=true,colorlinks,bookmarks=false]{hyperref}

\iccvfinalcopy % *** Uncomment this line for the final submission

\def\iccvPaperID{3794} % *** Enter the ICCV Paper ID here

\ificcvfinal\fi

\begin{document}

%%%%%%%%% TITLE
\title{FASA: Feature Augmentation and Sampling Adaptation \\ for Long-Tailed Instance Segmentation}

\author{
{\large
Yuhang Zang$^{1}$ ~~Chen Huang$^{2}$ ~~Chen Change Loy$^{1}$\textsuperscript{\Letter}} 
\\
{\large
${^1}$S-Lab, Nanyang Technological University ~~${^2}$Carnegie Mellon University
}
\\
{\tt\small $\{$zang0012, ccloy$\}$@ntu.edu.sg ~~chen-huang@apple.com}
}

\maketitle
\ificcvfinal\thispagestyle{empty}\fi
\input{sections/0_abstract.tex}
\input{sections/1_introduction.tex}
\input{sections/2_related_work.tex}
\input{sections/3_methodology.tex}
\input{sections/4_experiments.tex}
\input{sections/5_conclusion.tex}

{\small
\bibliographystyle{ieee_fullname}
\bibliography{sections/egbib}
}

\input{sections/6_appendix.tex}

\end{document}

%% file: sections/0_abstract.tex
% !TEX root = ../main.tex
\begin{abstract}
Recent methods for long-tailed instance segmentation still struggle on rare object classes with few training data. We propose a simple yet effective method, Feature Augmentation and Sampling Adaptation (FASA), that addresses the data scarcity issue by augmenting the feature space especially for rare classes. Both the Feature Augmentation (FA) and feature sampling components are adaptive to the actual training status --- FA is informed by the feature mean and variance of observed real samples from past iterations, and we sample the generated virtual features in a loss-adapted manner to avoid over-fitting. FASA does not require any elaborate loss design, and removes the need for inter-class transfer learning that often involves large cost and manually-defined head/tail class groups. We show FASA is a fast, generic method that can be easily plugged into standard or long-tailed segmentation frameworks, with consistent performance gains and little added cost. FASA is also applicable to other tasks like long-tailed classification with state-of-the-art performance.
\footnote{GitHub: \href{https://github.com/yuhangzang/FASA}{https://github.com/yuhangzang/FASA}.}
\footnote{Project page: \href{https://www.mmlab-ntu.com/project/fasa/index.html}{https://www.mmlab-ntu.com/project/fasa/index.html}.}
\end{abstract}

%% file: sections/1_introduction.tex
% !TEX root = ../main.tex
\section{Introduction}

\begin{figure*}[!htb]
\begin{minipage}{0.33\textwidth}
\centering
\includegraphics[width=\linewidth]{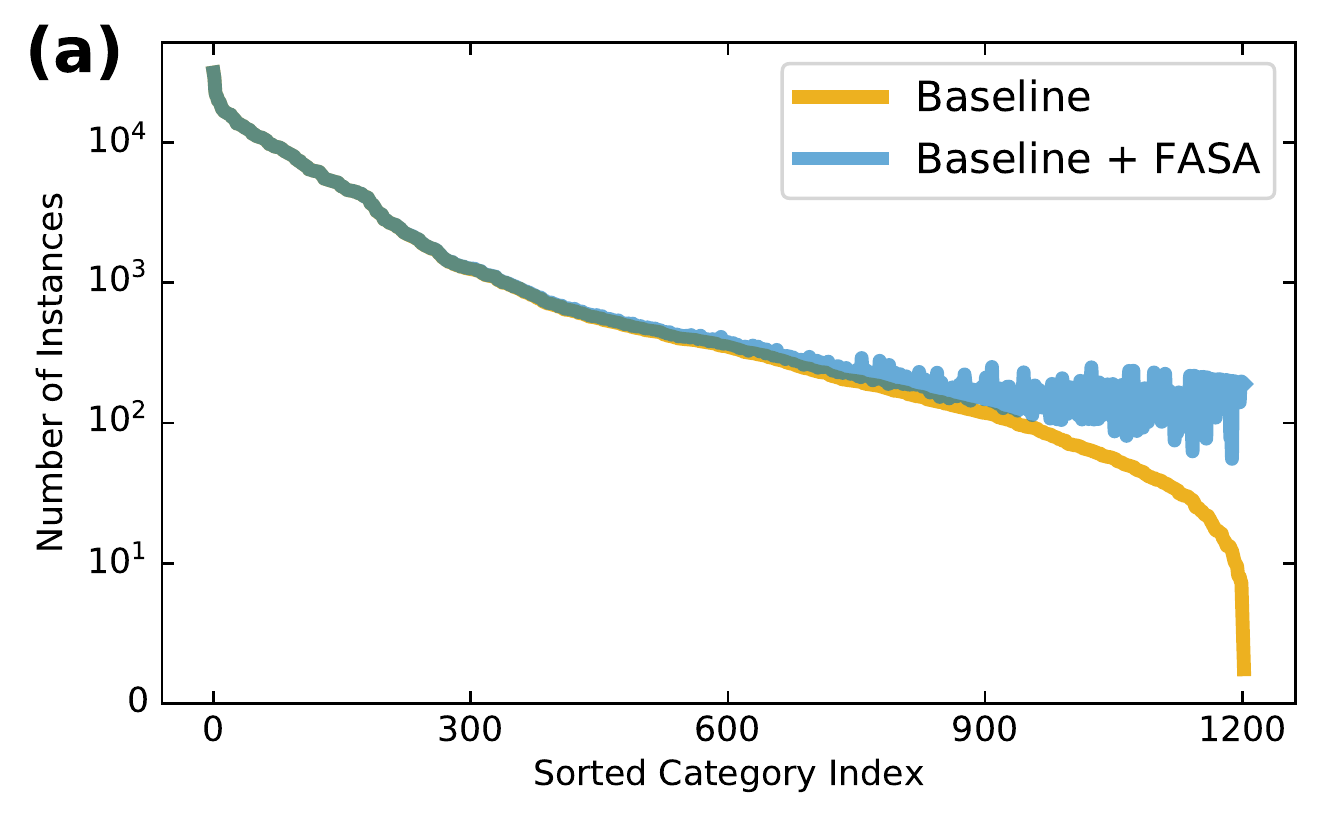}
\end{minipage}
\hfill
\begin{minipage}{0.33\textwidth}
\centering
\includegraphics[width=\linewidth]{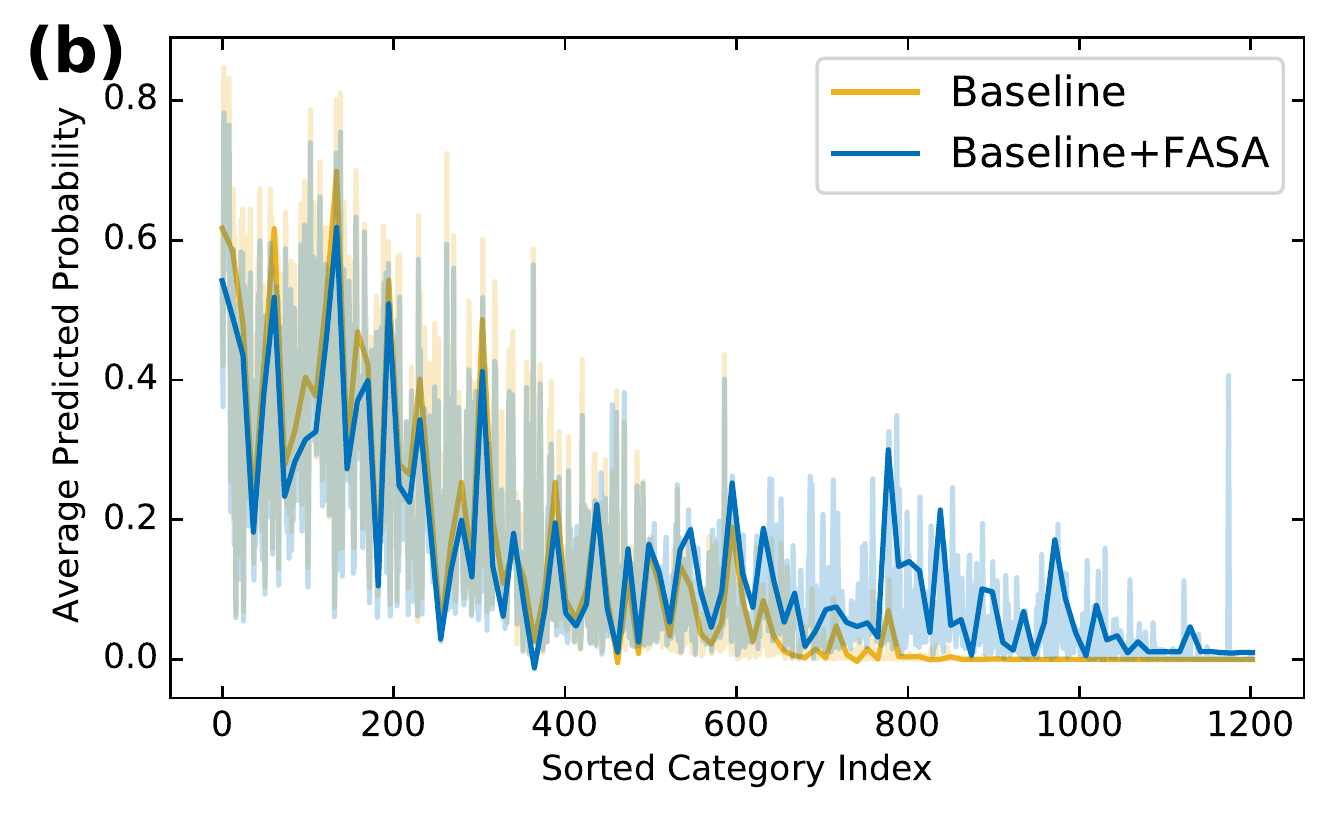}
\end{minipage}
\hfill
\begin{minipage}{0.33\textwidth}
\centering
\includegraphics[width=\linewidth]{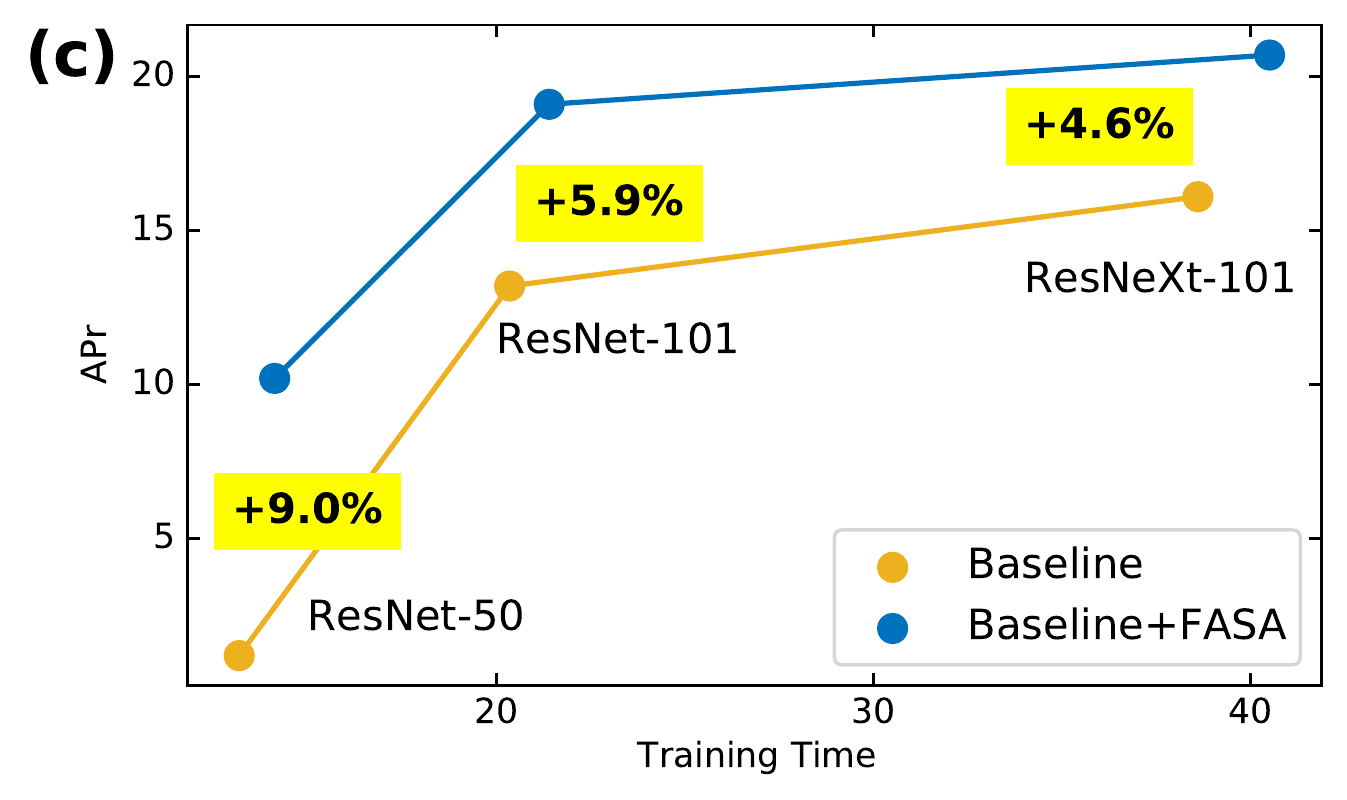}
\end{minipage}
\vspace{-6pt}
\caption{
\textbf{Class imbalance and the comparison of Mask R-CNN \cite{he2017mask} baseline with and without FASA on LVIS v1.0 dataset}. (a) By adaptive feature augmentation and sampling, our method FASA largely alleviates the imbalance issue, especially for rare classes.
(b) Compare the prediction results of FASA vs Mask R-CNN baseline regarding average category probability scores. The baseline model predicts near-zero scores for rare classes. While with FASA, rare-class scores are significantly boosted which merits final performance.
(c) FASA brings consistent improvements over different backbone models in mask \apr defined on rare classes. Such gains come at a very low cost (training time only increases by $\sim 3\%$ on average).
}
\label{fig:introduction}
\vspace{-14pt}
\end{figure*}

A growing number of methods are proposed to learn from long-tailed data in vision tasks like face recognition~\cite{huang2019deep}, image classification~\cite{liu2019large} and instance segmentation~\cite{gupta2019lvis}. We focus on the problem of long-tailed instance segmentation, which is particularly challenging due to the class imbalance issue. State-of-the-art methods~\cite{cai2019cascade,chen2019hybrid,he2017mask,lin2014microsoft} often fail to handle this issue and witness a large performance drop on rare object classes. Fig.~\ref{fig:introduction} exemplifies the struggle of the competitive Mask R-CNN~\cite{he2017mask} baseline on highly imbalanced LVIS v1.0 dataset~\cite{gupta2019lvis}.
We can see that there are about 300 rare classes in the tail that have no more than $10^2$ positive object instances. Such scarce training data result in poor performance for most of the tail classes, with near-zero class probabilities predicted for them.

To address the data scarcity issue, one intuitive option is to over-sample images that contain the tail-class objects~\cite{gupta2019lvis}. But the downside is that the over-sampled images will include more head-class objects at the same time due to the class co-occurrence within images. Hence for the instance segmentation task, re-sampling at the instance level is more desirable than at image level. Another option is data augmentation for considered objects, either in image space (\eg,~random flipping) or in feature space (\ie,~feature augmentation, on object regional features). Along this line, many methods have proven effective by augmenting the image/feature space of rare classes. The feature augmentation methods already show benefits for face recognition~\cite{Yin_2019_CVPR,liu2020deep}, person Re-ID~\cite{liu2020deep} and classification~\cite{kim2020m2m,chu2020feature}.
However, these methods require manual design of class groups, \eg, using heuristics like class size. And they often involve two stages of feature training and feature transfer, which induce additional cost.

Feature augmentation remains little explored for long-tailed instance segmentation tasks.
In this paper, we present an efficient and effective method, called Feature Augmentation and Sampling Adaptation (FASA). 
%We show that our feature augmentation and re-sampling schemes, when adapted to training status, suffice for long-tailed instance segmentation. 
%
FASA does not require any elaborate transfer learning or loss design~\cite{li2020overcoming,tan2020equalization,wu2020forest}. As a result, FASA keeps its simplicity, has extremely low complexity, while remains highly adaptive during training.
In the proposed method, we perform online Feature Augmentation (FA) for each class. The FA module generates class-wise virtual features using a distributional prior with its statistics calculated from previously observed real samples. This allows FA to capture class distributions and adapts to evolving feature space. 

\if 0
In the proposed method, we perform online Feature Augmentation (FA) for each class. The FA module generates class-wise virtual features using a distributional prior with its statistics calculated from previously observed real samples. This allows FA captures class distributions accurately, adapts to evolving feature space, and avoids generating irrelevant features. However, the augmented features can still be limiting for rare classes. 
%
This is because the \textit{observed} feature variance is often small with few training examples. Then the Gaussian ball computed for rare classes is likely incapable of sufficient FA. 
Although manipulating the Gaussian ball to generate more diverse virtual features is possible, human-designed manipulation schemes lack optimality guarantees, and transferring feature variances from head classes as in~\cite{chu2020feature,liu2020deep,Yin_2019_CVPR} may introduce harmful features that interfere with tail-class features.
\fi

The augmented features can still be limiting for rare classes. This is because the observed feature variance is often small with few training examples.
To overcome this problem, we adapt the sampling density of virtual features for each class. This way, our synthesized virtual features still live in the true feature manifold, just with different sampling probabilities to avoid under-fitting or over-fitting. We propose an adaptive sampling method here: when augmented features improve the corresponding class performance in validation loss, the feature sampling probabilities get increased, otherwise decreased. Such a loss-optimized sampling method is effective to re-balance the model predicting performance, see Fig~\ref{fig:introduction}(b). 

It is noteworthy that our full FASA approach handles virtual features only, thus it can serve as a plug-and-play module applicable to existing methods that learn real data with any re-sampling schemes or loss functions (standard or tailored for class imbalance). Comprehensive experiments of instance segmentation on LVIS v1.0~\cite{gupta2019lvis} and COCO-LT~\cite{wang2020devil} datasets demonstrate FASA as a generic component that can provide consistent improvement to other methods. Take LVIS dataset as an example. FASA improves Mask R-CNN~\cite{he2017mask} by $9.0\%$ and $3.3\%$ in mask AP metrics for rare and overall classes respectively, and improves a contemporary loss design~\cite{tan2020equalization} by $10.3\%$ and $2.3\%$. Moreover, these gains come at the cost of merely a $\sim 3\%$ increase in training time, see Fig.~\ref{fig:introduction}(c). Furthermore, FASA can generalize beyond the instance segmentation task, achieving state-of-the-art performance on long-tailed image classification as well. 

To summarize, the main \textbf{contributions} of this work is a fast and effective feature augmentation and sampling approach for long-tailed instance segmentation. The proposed FASA can be combined with existing methods in a plug-and-play manner. It achieves consistent gains and surpasses more sophisticatedly designed state-of-the-art methods. FASA also generalizes well to other long-tailed tasks.

%We summarize our \textbf{contributions} as follows:
%\begin{itemize}
%\setlength{\itemsep}{5pt}
%\setlength{\parsep}{0pt}
%\setlength{\parskip}{0pt}
%\item We propose a fast feature augmentation and sampling approach for long-tailed instance segmentation.
%\item Our approach FASA can be combined with existing methods in a plug-and-play manner. FASA achieves consistent gains and surpasses more sophisticated designed state-of-the-art methods with greater level of complexity.
%\item FASA generalizes well to other long-tailed tasks.
%\end{itemize}

% \cavan{I revised the intro so that it sounds less provoking when we discuss existing methods, and reduce unnecessary discussion.}

%% file: sections/2_related_work.tex
% !TEX root = ../main.tex
\section{Related Work}

\noindent \textbf{Long-tailed classification}.
For the long-tailed classification task, there is a rich body of widely used methods including data re-sampling~\cite{chawla2002smote} and re-weighting~\cite{cao2019learning,cui2019class}. Recent works~\cite{kang2019decoupling,zhou2020bbn} reveal the effectiveness of using different sampling schemes in decoupled training stages. Instance-balanced sampling is found useful for the first feature learning stage, which is followed by a classifier finetuning stage with class-balanced sampling.

\noindent \textbf{Long-tailed instance segmentation}. To handle class imbalance in instance segmentation task, recent methods still heavily rely on the ideas of data re-sampling~\cite{gupta2019lvis,shen2016relay,hu2020learning,wu2020forest}, re-weighting~\cite{tan2020equalization,ren2020balanced,tan2020equalization2,wang2020seesaw,zhou2020cn2,wang2021adaptive,zhang2021distribution} and decoupled training~\cite{li2020overcoming,wang2020devil}. For re-sampling, class-balanced sampling~\cite{shen2016relay} and Repeat Factor Sampling (RFS) \cite{gupta2019lvis} are conducted at the image level. However, image-level re-sampling sometimes worsens the imbalance at instance level due to the instance co-occurrence within images. The methods of data-balanced replay~\cite{hu2020learning} and NMS re-sampling~\cite{wu2020forest} belong to the category of instance-level re-sampling. For the re-weighting scheme, Equalization loss v1~\cite{tan2020equalization} and v2~\cite{tan2020equalization2} are such representative methods that re-weigh the sigmoid loss. 
More recent works~\cite{li2020overcoming,wu2020forest} attempt to divide imbalanced classes into relatively balanced class groups for robust learning. However, the class grouping process relies on static heuristics like class size or semantics which is not optimal. Tang~\etal~\cite{tang2020long} studied the sample co-existence effect in the long-tailed setting and proposed de-confounded training. Seesaw Loss~\cite{wang2020seesaw} dynamically rebalances the gradients of positive and negative samples, especially for rare categories.
Surprisingly, data augmentation as a simple technique, has been barely studied for long-tailed instance segmentation. In this paper, we show that competitive results can be obtained from smart data augmentation and re-sampling in the feature space, while being intuitively simple and computationally efficient. Also our approach is orthogonal to prior works, and can be easily combined with them to achieve consistent improvements.

\noindent \textbf{Data augmentation}. To avoid over-fitting and improve generalization, data augmentation is often used during network training. In the context of long-tailed recognition, data augmentation can also be used to supplement effective training data of under-represented rare classes, which helps to re-balance the performance across imbalanced classes.

There are two main categories of data augmentation methods: augmentation in the image space and feature space (\ie,~feature augmentation). Commonly used image-level augmentation methods include random image flipping, scaling, rotating and cropping. A number of advanced methods have also been proposed, such as Mixup~\cite{zhang2018mixup} and Cutmix~\cite{yun2019cutmix}. For our considered instance segmentation task, the copy-and-paste technique such as InstaBoost~\cite{fang2019instaboost} and Ghiasi~\etal~\cite{ghiasi2020simple} and prove effective. More recent works try to synthesize novel images using GAN~\cite{liu2019generative} or semi-supervised learning method~\cite{xie2020self}. Compared to the well established image-level data augmentation technique in deep learning, feature augmentation has not yet received enough attention.
On the other hand, Feature Augmentation (FA) directly manipulates the feature space, hence it can reshape the decision boundary for rare classes. Classic FA is built on SMOTE~\cite{chawla2002smote}-typed methods that interpolate neighboring feature points. More recently, researchers propose to Manifold Mixup~\cite{verma2019manifold} or MoEx~\cite{li2020feature} for better performance.
ISDA~\cite{wang2019implicit} augments data samples by translating CNN features along the semantically meaningful directions.
Such methods are often not directly applied for imbalanced class discrimination and suffer from high complexity.

% Liu~\etal~\cite{liu2020memory} considered the long-tailed setting and proposed to enhance the feature diversity of tail classes via a large memory of historical features. By contrast, our method only maintains compact feature mean and variance, and can use them to generate flexible virtual features.
% State-of-the-art FA methods benefit from feature transfer from head to tail classes,~\eg,~by transferring the features~\cite{chu2020feature,kim2020m2m}, feature variance~\cite{Yin_2019_CVPR} or angular variance~\cite{liu2020deep} between classes. However, these methods have notable limitations: may cause potential feature interference between classes, require predefined head/tail class groups and incur high cost in time and memory. Our approach waives the need for complex transfer learning and the associated costs. We compare with them in our experiments.
Some recent works have shown that feature augmentation benefits long-tail tasks such as face recognition~\cite{Yin_2019_CVPR}, person Re-ID~\cite{liu2020deep}, or long-tail classification~\cite{chu2020feature,kim2020m2m,liu2020memory,li2021metasaug}.
However, we observe that these methods have limitations when apply them to the long-tailed instance segmentation datasets such as LVIS~\cite{gupta2019lvis}.
Due to the high computational cost of instance segmentation task, some methods~\cite{Yin_2019_CVPR,kim2020m2m,chu2020feature} rely on two-phase pipeline or large memory of historical features\cite{liu2020memory} incur high cost in time and memory thus less efficient.
The special background class exists for the instance segmentation task (no class anchor), makes method~\cite{liu2020deep} that rely on margin-based classification loss less effective.
In addition, the small batch size of the instance segmentation frameworks limits the performance of approach~\cite{chu2020feature} that rely on mining confusing categories.
Our approach is specially designed for long-tailed instance segmentation, waives the need for complex two-phase training and the associated costs.
We compare with them in our experiments, see Sec~\ref{sec:compare_FA}.

%% file: sections/3_methodology.tex
% !TEX root = ../main.tex
\section{Methodology}

\begin{figure*}[ht]
\centering
\includegraphics[width=\linewidth]{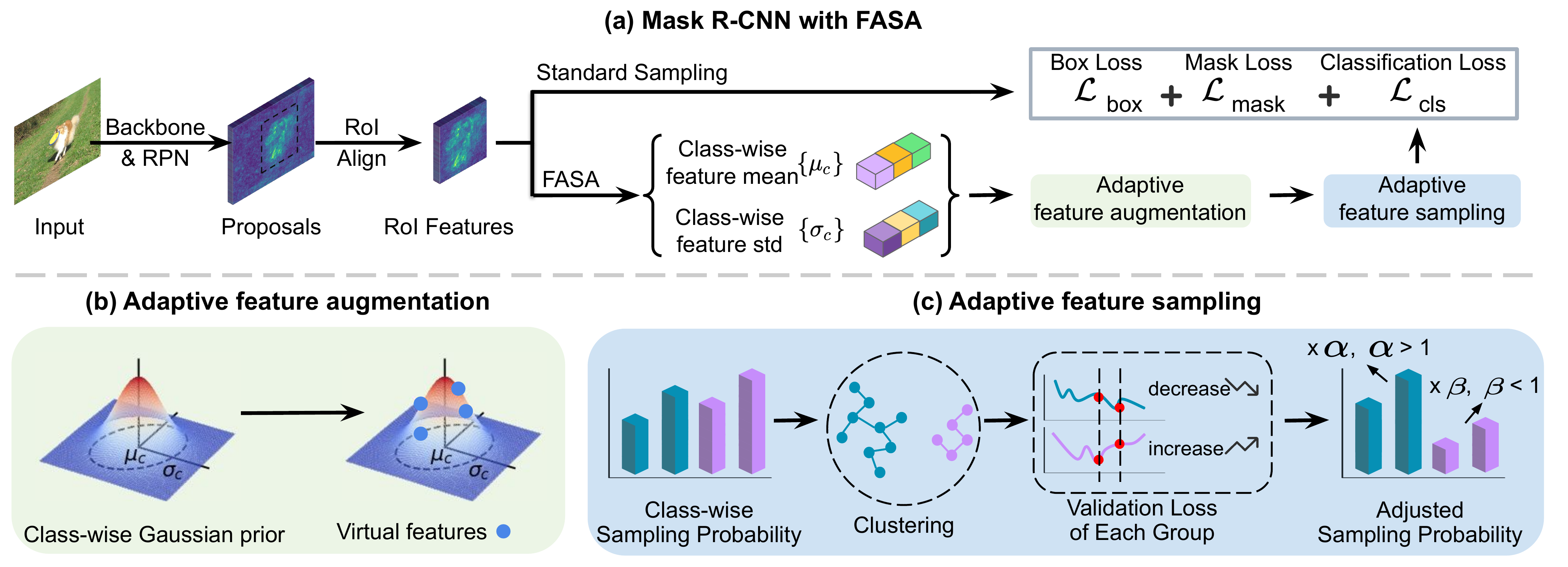}
\vspace{-12pt}
\caption{
(a) The pipeline of Mask R-CNN combined with the proposed FASA, a standalone module that generates virtual features to augment the classification branch for better performance on long-tailed data.
FASA maintains class-wise feature mean and variance online, followed by (b) adaptive feature augmentation and (c) adaptive feature sampling.
}
\label{fig:pipeline}
\vspace{-12pt}
\end{figure*}

In this section, we introduce the proposed Feature Augmentation and Sampling Adaptation (FASA) approach consisting of two components: 1) adaptive Feature Augmentation (FA) that generates virtual features to augment the feature space of all classes (especially for rare classes), 2) adaptive Feature Sampling (FS) that dynamically adjusts the sampling probability of virtual features for each class.

To better illustrate how FASA works for long-tailed instance segmentation, we take the Mask R-CNN~\cite{he2017mask} framework as our baseline and show an example of combining FASA and the segmentation baseline. The overall pipeline is shown in Fig.~\ref{fig:pipeline}. Note FASA is a standalone module for feature augmentation, leaving the baseline module unaltered. Therefore, FASA serves as a plug-and-play module and it can combine with and facilitate stronger baselines other than Mask R-CNN. We will show this flexibility in our experiments.

%Our experiments will demonstrate FASA's ability to improve the imbalanced learning performance on various baselines (architectures, advanced sampling and loss,~\etc).
 
Under the Mask R-CNN framework, the standard multi-task loss is defined for each Region of Interest (RoI):
\begin{equation}
    % \label{eq:loss_mask_rcnn}
    \mathcal{L}=\mathcal{L}_{cls}+\mathcal{L}_{box}+\mathcal{L}_{mask}.
    \label{eq1}
\end{equation}
For simplicity, we apply FASA to the classification branch only, which is the most vulnerable branch on long-tailed data as supported in Fig.~\ref{fig:introduction}(b). More discussion is provided in the supplementary materials. It is possible to augment other branches using FASA as well. We leave that to future work and expect further improvements.

\subsection{Adaptive Feature Augmentation}
\label{sec:virtual_sample}

Mask R-CNN is originally trained on ``real'' feature embeddings of positive region proposals generated by the Region Proposal Network (RPN). For each class, we aim to augment its real features that are likely scarce (\eg,~for rare classes). An ideal FA component should have the following properties: 1) generate diverse virtual features to enrich the feature space of corresponding class, 2) the generated virtual features capture intra-class variations accurately and do not deviate much from the true manifold which hinders learning, 3) adapt to the actual feature distribution that is evolving during training, 4) high efficiency.

To this end, we maintain an online Gaussian prior from previously observed real features, which fulfills the above-mentioned requirements. We found such prior suffices for FA purpose by generating up-to-date and diverse features even when the actual features are not Gaussian. Concretely, for each foreground class $c$ within the current batch, we can calculate the corresponding feature mean $\mu_c^t \in \mathbb{R}^d$ and standard deviation $\sigma_c^t  \in \mathbb{R}^d$ which together define a Gaussian feature distribution. Given the noisy nature of $\mu_c^t$ and $\sigma_c^t$, we use them to continuously update the more robust estimates $\mu_c$ and $\sigma_c$ instead, via a momentum mechanism:
\begin{equation}\begin{aligned}
\label{eq:virtual_sample}
\mu_{c} &\leftarrow (1 - m) \cdot =\mu_{c} + m \cdot \mu_{c}^t, \\
\sigma_{c} &\leftarrow (1 - m) \cdot \sigma_{c} + m \cdot \sigma_{c}^t, \\
 \end{aligned}\end{equation}
where $m$ is set to $0.1$ in all experiments. Then according to a Gaussian prior with up-to-date $\mu_{c}$ and $\sigma_{c}$, we generate the class-wise virtual features $\hat{x}_{c}$ by random perturbation under the feature independence assumption:
\begin{equation}
    \hat{x}_{c} = \mu_{c} + \sigma_{c} \odot \epsilon, \; \;\; \;   \epsilon \sim \mathcal{N}(0, I_d).
\end{equation}
The feature independence assumption allows for efficient FA, where augmented features are treated as IID random variables. We consider the covariance matrix of the Gaussian to be diagonal, thus largely reduce the complexity from $d^2$ to $d$.
Such generated virtual features $\{\hat{x}_{c}\}$ will need to be re-sampled (detailed later). Finally, both re-sampled $\{\hat{x}_{c}\}$ and real features $\{x_{c}\}$ (with its own sampling strategy in any baseline method) are sent to compute $\mathcal{L}_{cls}$ in Eq.~\eqref{eq1}.

\subsection{Adaptive Feature Sampling}
\label{sec:adaptive_sampling}

For a wise use of the generated virtual features for each class, we propose an adaptive Feature Sampling (FS) scheme to effectively avoid under-fitting or over-fitting from FA. The sampling process operates in a \textit{relative} fashion: if virtual features improve the corresponding class performance, their sampling probabilities get increased, otherwise decreased. Such relative adjustments of virtual feature sampling probabilities cater to the changing needs of FA during training. We can imagine that FA may be useful for rare classes during some training periods, but other training stages may call for decreased amount of FA to avoid over-fitting. By contrast, a static predefined sampling distribution would be independent of training dynamics and thus suboptimal.

\noindent\textbf{Parametric sampling formulation}. Note we still need an initial feature sampling distribution so as to adjust it later on. Obviously, we can benefit from a good initialization that avoids initial biased FA and costly adjustments. We choose to initialize the sampling distribution simply based on the inverse class frequency. It favors rare classes with higher sampling probabilities $p_c$ and is useful in our class-imbalanced setting with minimal assumption about data distribution. Now we have a predefined skewed sampling distribution at the beginning. Then we dynamically scale the per-class sampling probability $p_c$ as follows:
\begin{equation}
\label{eq:hard_mining}
    p_{c} = s_{c} \cdot \left(\frac{1}{N_{c}}\right) / \sum_{c=1}^C\left( \frac{1}{N_{c}} \right),
\end{equation}
where $s_c$ is the scaling factor to be estimated at each time, and $N_c$ denotes the class size of class $c$.

\noindent\textbf{Adaptive sampling scheme}. Recall the insight behind adapting the sampling probability $p_c$ is that: if FA does improve the performance of class $c$, then we should generate more virtual features and increase $p_c$; if we observe worse performance from FA, then $p_c$ is probably big enough and we should decrease it to avoid over-fitting with the augmented features. In practice, we use multiplicative adjustments to update $p_c$ every epoch. Specifically, we increase $p_{c}$ to $\min(1, p_{c} \cdot \alpha)$ with $\alpha = 1.1$, and decrease $p_{c}$ to $\max(0, p_{c} \cdot \beta)$ with $\beta = 0.9$.

One can adopt different performance metrics to guide the adjustment of $p_c$. Existing adaptive learning systems are either based on the surrogate loss~\cite{shrivastava2016training} or the more ideal actual evaluation metric~\cite{ALA_2019}. However for large-scale instance segmentation tasks, frequent evaluation of metrics like mAP on large datasets is very expensive. Therefore, we use the classification loss $\mathcal{L}_{cls}$ on validation set to adjust $p_c$ for our large-scale segmentation task. 
For the validation set, we apply Repeat Factor Sampling (RFS)~\cite{gupta2019lvis} method to balance the class distribution and provide more meaningful losses. 
The aforementioned setting works reasonably well in our experiments.

%This leads to a loss-adapted sampling scheme for virtual features which works reasonably well in our experiments. For lighter tasks or more efficient evaluation metrics, we can easily switch to the metrics as a better supervision signal.

\noindent\textbf{Group-wise adaptation}.
To make our method more robust to different scenarios, we need to address two common challenges in the long-tailed instance segmentation task. 
First, the per-class loss can be extremely noisy on limited evaluation data,~\eg,~for rare classes. In this case, both loss and metric cannot serve as reliable performance indicators. Second, the evaluation data may simply not be available for all classes. For example, the validation set of LVIS dataset~\cite{gupta2019lvis} only has 871 classes out of 1203 from training set. This makes it impossible to evaluate loss for the other 332 classes on validation set.

To solve the aforementioned issues, we propose to cluster all training classes into super-groups. Then we compute the average validation loss of within-group classes, and adjust their feature sampling probabilities together. In other words, we adjust the sampling probabilities by a single scaling factor ($\alpha$ or $\beta$) depending on the average of group-wise loss. By doing so, those classes with missing evaluation data can be safely ignored when computing the loss average, but their sampling probabilities can still be updated along with other classes within the same group. Moreover, the group-wise update is less noisy since it is based on the loss average computed on bigger data (from multiple classes).

For clustering, rather than using predefined heuristics like class size or semantics as in~\cite{li2020overcoming,wu2020forest}, we use the online class-wise feature mean $\mu_c$ and standard deviation $\sigma_c$. We adopt the density-based \cite{ester1996density} clustering algorithm using the following distance based on Fisher's ratio:
\begin{equation}
   d_{ij} = \frac{(\mu_{i} - \mu_{j})^{2}}{\sigma^{2}_{i} + \sigma^{2}_{j}}.
\end{equation}

The resulting super-groups are much more adaptive and meaningful than the predefined ones (\eg,~rare, common, and frequent class groups), and facilitate better group-wise feature re-sampling.
Please refer to the supplementary materials for the visualization of some super-groups of semantically similar classes.

\if 0
\noindent\textbf{Resampling of validation data}. 
Note a proper distribution of validation data is important to compute the validation loss as a good supervision signal for training. Ideally, the validation set should be class-balanced to provide unbiased supervision. However for long-tailed datasets like LVIS~\cite{gupta2019lvis}, the validation set is naturally imbalanced. Our experiments show that a re-sampled validation set provides a much better supervision than the original imbalanced one. Please refer to section \ref{sec:ablation} for a detailed comparison of re-sampling techniques and their impact to performance.
\fi

%% file: sections/4_experiments.tex
% !TEX root = ../main.tex
\section{Experiments}

\begin{table}[t]
    \tablestyle{8pt}{1.1}
	\caption{\small{Ablation study of the proposed FASA method on LVIS v1.0 \textit{validation} set. `FA' denotes adaptive Feature Augmentation, and `FS' denotes adaptive Feature Sampling. The metrics AP, \apr, \apc and \apf denote the mask mAP for overall, rare, common and frequent class groups.}}
    \label{tab:ablation_components}
	\centering
	\vspace{-6pt}
    \begin{tabular}{rrcx{30}cc}
    FA & FS & AP & \apr & \apc & \apf \\
    \midrule
    \gxmark & \gxmark & 20.8  & 8.0  & 20.2 & 27.0 \\
    \cmark & \gxmark & 22.3 \td{+1.5} & 12.7 \td{+5.7} & 21.7& 26.9 \\
    \cmark & \cmark & \bf 23.7 \bf{\td{+2.9}} & \bf 17.8 \bf{\td{+9.8}} & \bf 22.9 & \bf 27.2 \\
    \end{tabular}
	\vspace{-12pt}
\end{table}

\noindent\textbf{Datasets.}
Our experiments are conducted on two datasets: LVIS v1.0~\cite{gupta2019lvis} with 1203 categories and COCO-LT \cite{wang2020devil} with 80 categories.
Both of them are designed for long-tailed instance segmentation with highly class-imbalanced distribution.
We choose LVIS v1.0 dataset over LVIS v0.5 since the former has more labeled data for meaningful evaluations and comparisons.
LVIS v1.0 dataset defines three class groups: rare [1, 10), common [10, 100), and frequent [100, -) based on the number of images that contain at least one instance of the corresponding class.
Similarly, COCO-LT dataset defines four class groups [1, 20), [20, 400), [400, 8000), [8000, -).
We use the standard mean average precision (mAP) as the evaluation metric.
Using this metric on different class groups can well characterize the long-tailed class performance. Following~\cite{gupta2019lvis}, we denote the corresponding performance metrics of rare, common and frequent class groups as \apr, \apc and \apf.
Recently, Dave~\etal~\cite{dave2021evaluating} proposed that the mAP metric is sensitive to changes in cross-category ranking, and introduced two complementary metrics \fixedap and \appool. We also report the performance of FASA under the \fixedap and \appool metrics in the supplementary materials.

\noindent\textbf{Implementation details.}
Our implementation is based on the MMDetection~\cite{mmdetection} toolbox. We follow the same experimental setting as in~\cite{gupta2019lvis,tan2020equalization} for fair comparison. Please refer to supplementary materials for other details.
% The Mask R-CNN backbone is the ImageNet pre-trained ResNet-50~\cite{he2016deep} with a FPN~\cite{lin2017feature} neck and a box head with two sibling fully connected layers for RoI classification and regression.
% We apply random horizontal image flipping and multi-scale jittering with the smaller image sizes (640, 672, 704, 736, 768, 800) in all experiments.
% All models are trained with standard SGD on 8 NVIDIA V100 GPUs.
%The model and code will be released.

\subsection{Ablation Study on LVIS} \label{sec:ablation}
We first perform ablation studies on the large-scale LVIS dataset. We report the validation performance to ablate the core modules of our FASA approach.

\noindent\textbf{Effectiveness of FA and FS.}
Table~\ref{tab:ablation_components} verifies the critical roles of our adaptive modules for Feature Augmentation (FA) and Feature Sampling (FS).
The baseline (top row) only performs repeat factor sampling~\cite{gupta2019lvis} for real features, without using any FASA components. Our FA module (second row) significantly improves the performance, both for overall and rare classes. Our adaptive FS further boosts the performance, especially in the rare class groups ($AP_{r}^{m}$ $12.7\% \rightarrow 17.8\%$) with other groups being competitive. The results suggest the effectiveness of both the FA and FS components in improving the training performance.

\begin{table}[t]
    \tablestyle{7pt}{1.1}
    \caption{\small{Comparing our FASA with 1) \textbf{interpolation} based methods SMOTE~\cite{chawla2002smote} and MoEx~\cite{li2020feature}, 2) \textbf{copy-and-paste} base methods InstaBoost~\cite{fang2019instaboost}, 3) \textbf{feature augmentation} based methods Yin~\etal~\cite{Yin_2019_CVPR}, Liu~\etal~\cite{liu2020deep}, Chu~\etal~\cite{chu2020feature} on LVIS v1.0 \textit{validation} set. The baseline (top row) denotes Mask R-CNN~\cite{he2017mask} without any augmentation.}}
    \label{tab:ablation_augmentation}
	\vspace{-6pt}
	\centering
	\begin{tabular}{ccx{30}cc}
        Augmentation  & AP & \apr & \apc & \apf \\
        \midrule
        \gxmark & 20.8 & 8.0 & 20.2 & 27.0 \\
        \midrule
        SMOTE~\cite{chawla2002smote} & 21.5 \td{+0.7} & 10.2 \td{+2.2} & 20.9  & 27.1 \\
        MoEx, CVPR'21~\cite{li2020feature} & 21.2 \td{+0.4} & 9.2 \td{+1.2} & 20.6 & 27.1 \\
        InstaBoost, CVPR'19~\cite{fang2019instaboost} & 21.4 \td{+0.6} & 10.3 \td{+2.3} & 20.7 & \bf 27.2 \\
        Yin~\etal, CVPR'19~\cite{Yin_2019_CVPR} & 21.6 \td{+0.8} & 11.1 \td{+3.1} & 20.9 & 27.1 \\
        Liu~\etal, CVPR'20~\cite{liu2020deep} & 21.0 \td{+0.2} & 9.6 \td{+1.6} & 20.1 & 26.8 \\
        Chu~\etal, ECCV'20~\cite{chu2020feature} & 21.4 \td{+0.6} & 9.7 \td{+1.7} & 21.0 & 27.0 \\
        \midrule
        FASA (ours) & \bf 23.7 \bf{\td{+2.9}} & \bf 17.8 \bf{\td{+9.8}} & \bf 22.9 & \bf 27.2 \\
        \end{tabular}
	\vspace{-12pt}
\end{table}

\noindent\textbf{Comparison with other augmentation methods.}
\label{sec:compare_FA}
To further show the simplicity and efficacy of our method, we compare with the classic SMOTE method~\cite{chawla2002smote}, MoEx~\cite{li2020feature}, InstaBoost~\cite{fang2019instaboost} and state-of-the-art methods~\cite{liu2020deep,chu2020feature,Yin_2019_CVPR} that are specially designed for the long-tailed setting.
Since~\cite{liu2020deep,chu2020feature,Yin_2019_CVPR} only report results on face recognition and person Re-ID, while no public codes are available, we re-implemented them and optimized their parameters and performance for LVIS (see the supplementary materials for implementation details).

Table~\ref{tab:ablation_augmentation} shows favorable results of our FASA when compared to others. Our gains are particularly apparent for \apr and \apc.
% Our \apf is a bit worse but can be boosted when combined with adaptive feature sampling in the following stage.
% The comparative results confirm the need of FA to adapt to the actual feature distribution, which is missing in prior works. 
% Our FA method achieves this by maintaining online the feature mean and variance per class. The involved operations are equally simple as SMOTE, and much faster than InstaBoost. Our FA method only incurs a small extra memory which is constant.
Since SMOTE~\cite{chawla2002smote}, MoEx~\cite{li2020feature} and InstaBoost~\cite{fang2019instaboost} are not directly designed for long-tailed setting, we verify that our FASA obtains a favorable performance over them. 
For more related feature augmentation approaches~\cite{chu2020feature,liu2020deep,Yin_2019_CVPR}, we observe some limitations when transferring them into the instance segmentation task.
Liu~\etal~\cite{liu2020deep} are based on margin-based face recognition loss such as ArcFace~\cite{deng2019arcface} that constrains the margin between each instance and its class anchor. 
Unfortunately, since the instance segmentation task has to deal with the special \textit{background} class that no distinct anchor, the margin-based loss does not perform well on LVIS.
In contrast, our approach is not restricted by the form of loss function.
As for Chu~\etal~\cite{chu2020feature}, the performance is limited by the small batch size of the instance segmentation task, which does not guarantee the selection of top confusing samples.
Both~\cite{chu2020feature} and~\cite{Yin_2019_CVPR} apply feature transfer from head to tail classes. They use a two-stage training pipeline that requires a pre-trained model to extract features.
In contrast, our proposed FA method can be trained end-to-end, much faster than ~\cite{chu2020feature,Yin_2019_CVPR} and only incurs a small extra memory which is constant.

% \cavan{Reference~\cite{li2020feature} is not discussed in the related work. We may want to add that. The paper has been accepted by CVPR 2021. Please update the citation. I also added the conference-year in Table~\ref{tab:ablation_augmentation}.}

\begin{table}[t]
    \tablestyle{8pt}{1.1}
	\caption{\small{Comparison between our adaptive feature sampling strategy and a static one (with fixed scaling factor $s_{c}$ in Eq.~\eqref{eq:hard_mining}).}
}
    \label{tab:ablation_sampling}
	\vspace{-6pt}
	\centering
    \begin{tabular}{lcccc}
    Sampling Method  & AP & \apr & \apc & \apf \\
    \midrule
    Static ($s$ = 1)  & 21.7 & 12.2 & 20.8 & 27.0 \\
    Static ($s$ = 5) & 22.3 & 12.7 & 21.7 & 26.9 \\
    Static ($s$ = 15)  & 21.3 & 12.0 & 20.2 & 26.6 \\
    \midrule
    %Adaptive (original val set) & 23.0 & 15.6 & 21.6 & 27.1 \\
    %Adaptive (re-sampled val set) & \textbf{23.7} & \textbf{17.8} & \textbf{22.9} & \textbf{27.2} \\
    Adaptive & \textbf{23.7} & \textbf{17.8} & \textbf{22.9} & \textbf{27.2} \\
    %Adaptive (balanced val subset) & 23.2 & 16.6 & 22.4 & 27.1 \\
    \end{tabular}
	\vspace{-12pt}
\end{table}

\noindent\textbf{Analysis of adaptive FS.}
Recall in Eq.~\eqref{eq:hard_mining}, our class-wise feature sampling probabilities are adaptively adjusted by the scaling factor $s_{c}$ in an adaptive manner. Table~\ref{tab:ablation_sampling} validates our design, showing that sampling is poor with a static sampling strategy.
Specifically, we observe that $s_{c}=5$ works best for \apr, but is not optimal for \apf that needs $s_{c}=1$ instead.
On the other hand, our adaptive FS tunes $s_c$ online to effectively re-balance the performance across categories.

% \cavan{I removed the part on trying different validation set settings. As has been shown in CVPR meta reviews from AC, this experiments draws further doubts on our method. Of course, we revise the methodology accordingly.}

% \cavan{I prefer the original version with the figure that describe the sampling behaviour. That is more informative. I put it back here.}

\begin{figure*}[t]
    \begin{minipage}{0.4\textwidth}
    \centering
    \includegraphics[width=0.9\textwidth]{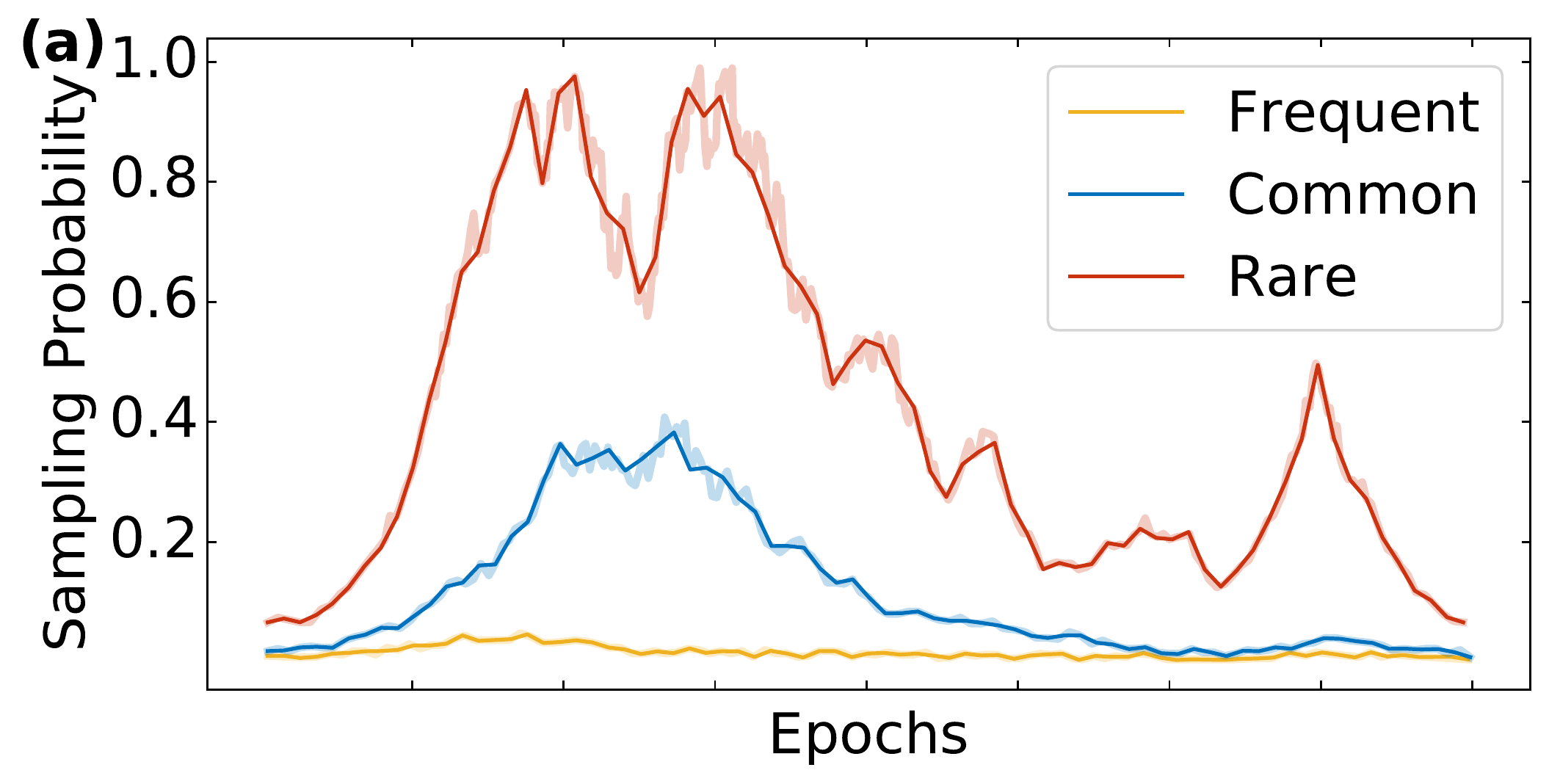}
    \end{minipage}
    \hfill
    \begin{minipage}{0.6\textwidth}
    \centering
    \includegraphics[width=1.0\textwidth]{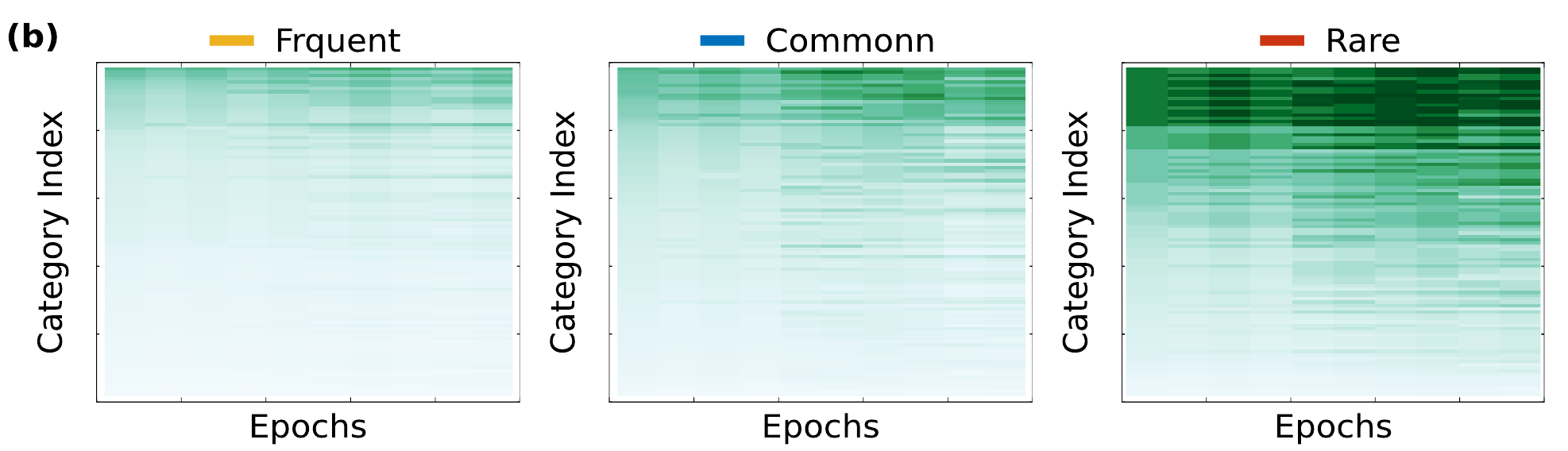}
    \end{minipage}
    \vspace{-12pt}
    \caption{
    \textbf{Visualization of the class-wise sampling probabilities during training}.
    (a) We show the average sampling probabilities for the rare/common/frequent class groups.
    (b) The changing behavior of class-wise sampling probabilities within each group. 
    }
    \label{fig:adaptive_sampling}
\end{figure*}

Fig.~\ref{fig:adaptive_sampling}(a) depicts how the class-wise sampling probabilities change during training.
Overall, the rare classes exhibit high sampling probabilities compared with the common and frequent classes.
The rare-class sampling probabilities typically get increased to use more virtual features at the beginning. Then they gradually decrease to avoid over-fitting. Upon convergence, when the learning rate decreases, the sampling probabilities of rare categories see a bump to help with the corresponding classifier 'fine-tuning'. By contrast, the changes of sampling probabilities for the common and frequent classes are smaller. Fig.~\ref{fig:adaptive_sampling}(b) further shows the dynamic changes of sampling probabilities within each class group.

\if 0
We also compared different ways of validation set re-sampling, which will provide different supervision signals for adaptive FS. Three methods are compared: 1) The original imbalanced validation set is used, without any re-sampling; 2) Applying the Repeat Factor Sampling (RFS)~\cite{gupta2019lvis} method to validation data for re-balancing purpose; 3) Selecting a relatively balanced subset from the validation set of LVIS v1.0. To do so, we simply removed images that do not contain rare or common classes, resulting a subset of 3286 images out of 19806. As shown in Table~\ref{tab:ablation_sampling}, it is indeed favorable to do re-sampling to validation set. Since the re-sampling approach RFS performs best, we use it as our default method.
\fi

\subsection{Comparison with State of the Arts on LVIS}
\begin{table*}[t]
    \centering
    \tablestyle{8pt}{1.0}
	\caption{\small{Comparing state-of-the-art methods with and without our FASA on the LVIS v1.0 \textit{validation} dataset. We compare with the Mask R-CNN baseline, and state-of-the-art re-sampling approach Repeat Factor Sampling (RFS)~\cite{gupta2019lvis}, Equalization Loss (EQL)~\cite{tan2020equalization}, Classifier Re-training (cRT)~\cite{kang2019decoupling}, Balanced Group Softmax (BAGS)~\cite{li2020overcoming} and Seesaw Loss~\cite{wang2020seesaw}. The `Uniform' method indicates random and uniform sample images. These methods are trained under different training schedules (24 or 12+12 epochs) using the public codes. All methods use ResNet-50~\cite{he2016deep} as the backbone for fair comparison.}}
	\label{tab:ablation_ce}
	\vspace{-6pt}

    \begin{tabular}{ccccx{20}x{30}cc}
    Loss & Sampler & $\#$Epochs & FASA & AP & \apr  & \apc & \apf \\
    \midrule
    \multirow{4}*{Softmax CE} & \multirow{2}*{Uniform} & \multirow{2}*{24} & \gxmark & 19.3 & 1.2 & 17.4 & \textbf{29.3} \\
    ~ & ~ & ~ & \cmark & \textbf{22.6} \bf{\td{+3.3}} & \textbf{10.2} \bf{\td{+9.0}} & \textbf{21.6}   & 29.2 \\
    \cline{2-8}
    ~ & \multirow{2}*{RFS, CVPR'19~\cite{gupta2019lvis}} &  \multirow{2}*{24} & \gxmark & 22.8 & 12.9 & 21.6 & 28.3 \\
    ~ & ~ & & \cmark &\textbf{24.1} \bf{\td{+1.3}} & \textbf{17.3} \bf{\td{+4.4}} & \textbf{22.9} & \textbf{28.5} \\
    \midrule
    \multirow{2}*{EQL, CVPR'20~\cite{tan2020equalization}} & \multirow{2}*{Uniform} & \multirow{2}*{24} & \gxmark     & 22.1 & 5.1 & 22.4 & 29.3 \\
    ~  & ~ & ~ & \cmark & \textbf{24.4} \bf{\td{+2.3}} & \textbf{15.4} \bf{\td{+10.3}} & \textbf{23.5} & \textbf{29.4} \\
    \midrule
    \multirow{2}*{cRT, ICLR'20~\cite{kang2019decoupling}} & \multirow{2}*{Uniform / RFS, CVPR'19~\cite{gupta2019lvis}} & \multirow{2}*{12+12}& \gxmark     & 22.4 & 12.2 & 20.4 & \textbf{29.1} \\
    ~  & ~ & ~ & \cmark &\textbf{23.6} \bf{\td{+1.2}} & \textbf{15.1} \bf{\td{+2.9}} & \textbf{22.0} & \textbf{29.1} \\
    
    \midrule
    \multirow{2}*{BAGS' CVPR'20~\cite{li2020overcoming}} & \multirow{2}*{Uniform / RFS, CVPR'19~\cite{gupta2019lvis}} & \multirow{2}*{12+12} & \gxmark & 22.8 & 12.4 & 22.2 & \textbf{28.3} \\
    ~ & ~ & ~ & \cmark & \textbf{24.0} \bf{\td{+1.2}} & \textbf{15.2} \bf{\td{+2.8}} & \textbf{23.4} & \textbf{28.3} \\
    
    \midrule
    \multirow{2}*{Seesaw, CVPR'21~\cite{wang2020seesaw}} & \multirow{2}*{RFS, CVPR'19~\cite{gupta2019lvis}} & \multirow{2}*{24} & \gxmark     & 26.4 & 19.6 & 26.1 & 29.8 \\
    ~  & ~ & ~ & \cmark & \textbf{27.5} \bf{\td{+1.1}} & \textbf{21.0} \bf{\td{+1.4}} & \textbf{27.5} & \textbf{30.1} \\
    \end{tabular}
 	\vspace{-12pt}
\end{table*}

\begin{table*}[t]
    \centering
    \tablestyle{5.6pt}{1.0}
	\caption{\small{Comparison state-of-the-art methods with and without FASA, using large backbones (ResNet-101 \cite{he2016deep}, ResNeXt-101-32x8d \cite{xie2017aggregated}) and advanced Cascade Mask R-CNN~\cite{cai2019cascade} framework.}}
	\label{tab:ablation_sota}
	\vspace{-6pt}
    \begin{tabular}{cccccx{20}x{30}cc}
    Method & Loss & Sampler & Backbone & FASA & AP & \apr & \apc & \apf \\
    \if 0
    \midrule
    \multirow{2}*{Mask R-CNN, ICCV'17~\cite{he2017mask}} & \multirow{2}*{R101} & \gxmark & 20.7 & 1.1 & 19.3 & \textbf{30.8} \\
    ~ & ~ & ~ & ~ & \cmark & \textbf{23.5} \bf{\td{+2.8}} & \textbf{8.2} \bf{\td{+7.1}} & \textbf{23.2} & \textbf{30.8} \\
    \fi
    \midrule
    \multirow{2}*{Mask R-CNN, ICCV'17~\cite{he2017mask}} & \multirow{2}*{Softmax CE} & \multirow{2}*{RFS, CVPR'19~\cite{gupta2019lvis}} & \multirow{2}*{R101} & \gxmark & 24.4 & 13.2 & 24.7 & 30.3 \\
    ~ & ~ & ~ & ~ & \cmark & \textbf{26.3} \bf{\td{+1.9}} & \textbf{19.1} \bf{\td{+5.9}} & \textbf{25.4} & \textbf{30.6} \\
    
    \midrule
    \multirow{2}*{Mask R-CNN, ICCV'17~\cite{he2017mask}} & \multirow{2}*{Softmax CE} & \multirow{2}*{RFS, CVPR'19~\cite{gupta2019lvis}} & \multirow{2}*{X101} & \gxmark & 26.1 & 16.1 & 24.9 & \textbf{32.0} \\
    ~ & ~ & ~ & ~ & \cmark & \textbf{27.7} \bf{\td{+1.6}} & \textbf{20.7} \bf{\td{+4.6}} & \textbf{26.6} & \textbf{32.0} \\
    \midrule
    \multirow{2}*{Cascade Mask R-CNN, TPAMI'19~\cite{cai2019cascade}} & \multirow{2}*{Softmax CE} & \multirow{2}*{RFS, CVPR'19~\cite{gupta2019lvis}} & \multirow{2}*{R101} & \gxmark & 25.4 & 13.7 & 24.8 & 31.4 \\
    ~ & ~ & ~ & ~ & \cmark & \textbf{27.7} \bf{\td{+2.3}} & \textbf{19.8} \bf{\td{+5.9}} & \textbf{27.3} & \textbf{31.6} \\
    \midrule
    \multirow{2}*{Cascade Mask R-CNN, TPAMI'19~\cite{cai2019cascade}} & \multirow{2}*{Seesaw, CVPR'21~\cite{wang2020seesaw}} & \multirow{2}*{RFS, CVPR'19~\cite{gupta2019lvis}} & \multirow{2}*{R101} & \gxmark & 30.1 & 21.4 & 30.0 & 33.9 \\
    ~ & ~ & ~ & ~ & \cmark & \textbf{31.5} \bf{\td{+1.4}} & \textbf{24.1} \bf{\td{+2.7}} & \textbf{31.9} & \textbf{34.0} \\
    \end{tabular}
	\vspace{-12pt}
\end{table*}

In this section, we evaluate the full FASA method against state-of-the-art methods on the LVIS v1.0 dataset.
We consider the following representative methods:
1) \textit{Repeat Factor Sampling (RFS)} \cite{gupta2019lvis} is an image-level data re-sampling technique.
    We use the same repeat factor $1e-3$ as proposed in the original paper.
2) \textit{Equalization Loss (EQL)}~\cite{tan2020equalization} is a loss re-weighting method that aims to ignore the harmful gradients from rare classes.
3) \textit{Classifier Re-training (cRT)}~\cite{kang2019decoupling} first uses random sampling for feature representation learning, and then re-trains the classifier with repeat factor sampling.
4) \textit{Balanced Group Softmax (BAGS)}~\cite{li2020overcoming} first performs class grouping and then makes the classification loss relatively balanced across those class groups. Grouping simply relies on the class sizes that are independent of training dynamics and suboptimal.
5) \textit{Seesaw Loss~\cite{wang2020seesaw}} balances ratio of cumulative gradients for the positive samples and negative samples of different categories.

\noindent\textbf{Comparing using Mask R-CNN baseline.}
The aforementioned methods have shown solid improvements when combined with the Mask R-CNN baseline. Here we conduct a more comprehensive comparison with even stronger versions of these methods under different training schedules. Specifically, we experimented with the default 12 and 24 epochs schedules~\cite{mmdetection}, as well as the decoupled two-stage training schedule~\cite{kang2019decoupling,li2020overcoming}.
In the first stage, we train the model for 12 epochs with standard random data sampling and cross-entropy loss. Then in the second stage, we fine-tune for 12 epochs using those advanced re-sampling or re-weight approaches, such as RFS and BAGS.
We refer to this schedule as `12+12'.
% Note the BAGS method uses such decoupled training by default thus has very strong results.
We compare with all of these methods by plugging in our FASA module and comparing the performance difference.

Table~\ref{tab:ablation_ce} summarizes the comparison results. Specifically, we repeat each experiment three times with different random seeds and report the mean value of results.
% When we combine our FASA with vanilla Mask R-CNN, we observe huge gains in the overall metric AP - 4.9\% for 12 epochs and 3.3\% for 24 epochs.
When we combine our FASA with vanilla Mask R-CNN, we observe huge 3.3\% gains in the overall metric AP. Our benefits are particularly evident in the rare class performance \apr, with gains of 9.0\%. This validates FASA's superior ability to deal with long-tailed tasks. We also verify that the benefits of FASA remain consistent over multiple runs. When combined with stronger methods (RFS/EQL/cRT/BAGS/Seesaw) or training schedules, FASA can still obtain consistent improvements in overall AP, where the rare-class improvements in \apr dominate. Such gains come without compromising the common- and frequent-class metrics \apc and \apf, where FASA performs better or remains on par.

\noindent\textbf{Long training schedules and large backbones.}
FASA works well even with large backbones (ResNet 101~\cite{he2016deep}, ResNeXt 101-32-8d~\cite{xie2017aggregated}) and advanced instance segmentation framework Cascade Mask R-CNN~\cite{cai2019cascade}.
As shown in Table~\ref{tab:ablation_sota}, the benefits of FASA still hold under different settings. On the validation set, FASA improves the performance of Mask R-CNN baseline (ResNet-101 backbone) by a large margin. We observe improvements in overall AP and rare-class \apr by 1.9\% and 5.9\% respectively.
For the stronger ResNeXt101 backbone, we observe similar trend.
FASA significantly improves the rare class performance \apr by 4.6\%.
For Cascade Mask R-CNN~\cite{cai2019cascade} framework and Seesaw loss~\cite{wang2020seesaw}, FASA offers consistent performance boost in rare classes while remaining strong on both common and frequent classes.

\begin{figure}[t]
    \centering
    \includegraphics[width=\linewidth]{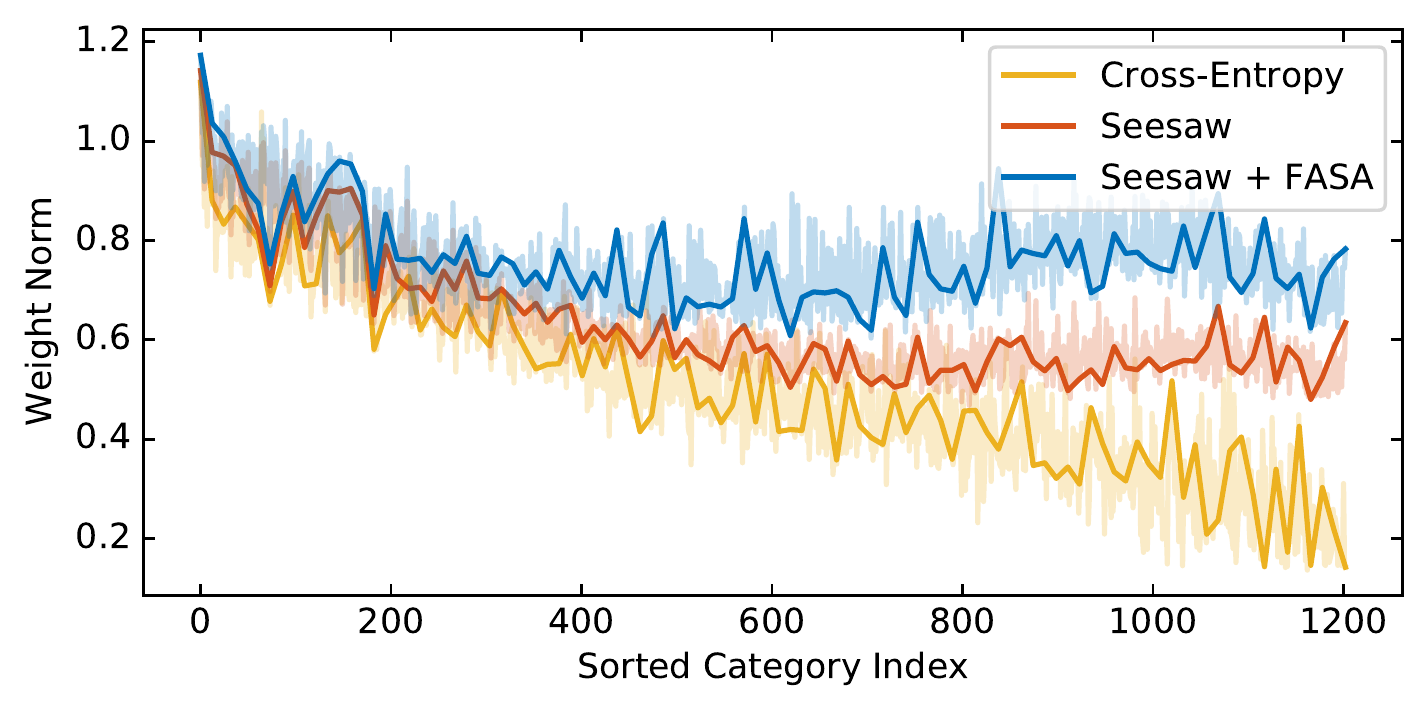}
	\vspace{-6pt}
	\caption{\small{Comparison of the classifier weight norm of Cross-Entropy and Seesaw Loss~\cite{wang2020seesaw} (with and without FASA). Our FASA leads to a more class-balanced weight norm distribution.}}
	\label{fig:weight_norm}
	\vspace{-12pt}
\end{figure}

\noindent\textbf{Analyzing classifier weight norm.} As discussed in~\cite{kang2019decoupling,li2020overcoming}, the weight norm of a classifier is correlated with the imbalanced learning performance, with the weight norms of tail classes often being much smaller on long-tailed data.
Fig.~\ref{fig:weight_norm} visualizes the weight norm of classifiers trained with and without FASA.
It can be seen that FASA leads to more balanced class distributions of weight norms than Cross-Entropy baseline and Seesaw Loss~\cite{wang2020seesaw}.
This partially explains why FASA greatly improves the rare-class performance of these methods.

\subsection{Evaluation on COCO-LT}
We evaluated FASA on COCO-LT~\cite{wang2020devil} dataset to examine the generalizability of our approach.
For a fair comparison, we follow the same experiment setting as SimCal~\cite{wang2020devil}.
There are two main differences compared to the implementation details we described in the LVIS dataset:
(1) Single-scale training is used.
We set the input image's short side to 800 pixels and do not use multi-scale jittering.
(2) The bounding box head and mask head are class agnostic.

Table~\ref{tab:cocolt} summarizes the results.
The top row shows results of the Mask R-CNN~\cite{he2017mask} with ResNet50~\cite{he2016deep} backbone and FPN~\cite{lin2017feature} neck.
The second row shows results obtained by the previous state-of-the-art method SimCal \cite{wang2020devil}, which involves classification calibration training and then dual head inference.
The third row shows results of the Mask R-CNN baseline augmented with our FASA.
Note how FASA improves the performance of Mask R-CNN considerably, especially for those rare classes ($\mathrm{AP}^{\mathrm{m}}_{\mathrm{1}}$ and $\mathrm{AP}^{\mathrm{m}}_{\mathrm{2}}$).
Overall, FASA performs better than SimCal \cite{wang2020devil} without any form of decoupled training. 

\begin{table}[t]
    \centering
    \tablestyle{5pt}{1.1}
    \caption{
    \small{Results on COCO-LT \cite{wang2020devil} \textit{minival} set. $\mathrm{AP}^{\mathrm{m}}$ and $\mathrm{AP}^{\mathrm{b}}$ denote the Mask mAP and Bbox mAP, respectively. $\mathrm{AP}^{\mathrm{m}}_{\mathrm{1}}$, $\mathrm{AP}^{\mathrm{m}}_{\mathrm{2}}$, $\mathrm{AP}^{\mathrm{m}}_{\mathrm{3}}$, $\mathrm{AP}^{\mathrm{m}}_{\mathrm{4}}$ refer to bin of $[1, 20)$, $[20, 400)$, $[400, 8000)$, $[8000, -)$ training instances.}}
    \label{tab:cocolt}
	\vspace{-6pt}
    \begin{tabular}{ccccccc}
    Method & $\mathrm{AP}^{\mathrm{m}}$ & $\mathrm{AP}^{\mathrm{m}}_{\mathrm{1}}$ & $\mathrm{AP}^{\mathrm{m}}_{\mathrm{2}}$ & $\mathrm{AP}^{\mathrm{m}}_{\mathrm{3}}$ & $\mathrm{AP}^{\mathrm{m}}_{\mathrm{4}}$ & $\mathrm{AP}^{\mathrm{b}}$ \\
    \midrule
    Mask R-CNN & 18.7 & 0.0 & 8.2 & 24.4 & 26.0 & 21.4 \\
    SimCal \cite{wang2020devil} & 21.8 & \textbf{15.0}         & 16.2         & 24.3         & 26.0         & 24.6 \\
    \midrule
    FASA (Ours)    & \textbf{23.4}        & 13.5         & \textbf{19.0}         & \textbf{25.2}         & \textbf{27.5}         & \textbf{26.0} \\
    \end{tabular}
	\vspace{-12pt}
\end{table}

\subsection{Evaluation on CIFAR-LT-100}
\begin{table}[t]
    \centering
    \tablestyle{4pt}{1.1}
    \caption{\small{Comparison of classification accuracy on CIFAR-LT-100 \cite{cao2019learning} dataset with imbalance ratio 100. $\dagger$ denotes that the results are copied from \cite{kim2020m2m}.}}
    \label{tab:cifar100lt}
    \vspace{-6pt}
    \begin{tabular}{cccc}
    Backnone & Method & Augmentation & Accuracy ($\%$) \\
    \midrule
    \multirow{5}*{ResNet 32} & LDAM \cite{cao2019learning} $\dagger$ & \xmark & 42.0 $\pm$ 0.09 \\
    ~ & LDAM \cite{cao2019learning} & M2M \cite{kim2020m2m} & 43.5 $\pm$ 0.22 \\
    ~ & LDAM \cite{cao2019learning} & FASA(Ours)            & \textbf{43.7 $\pm$ 0.25} \\
    \cline{2-4}
    ~ & De-confound-TDE~\cite{tang2020long} & \xmark       & 44.1 \\
    ~ & De-confound-TDE~\cite{tang2020long} & FASA (Ours)   & \textbf{45.2 $\pm$ 0.16} \\
    \midrule
    \multirow{2}*{ResNet 34} & Cross entropy & Chu~\etal~\cite{chu2020feature} & 48.5 \\
    ~ & Cross entropy & FASA(Ours)            & \textbf{49.1 $\pm$ 0.23} \\
    \end{tabular}
	\vspace{-12pt}
\end{table}
To demonstrate the generalizability of FASA in other domains, we further apply FASA on the long-tailed image classification task.
We follow \cite{cao2019learning} to conduct experiments on the long-tailed CIFAR-100 dataset with a substantial imbalance ratio of $\rho=100$ (the ratio between sample sizes of the most frequent and least frequent class, $\rho=\max_i\{N_i\}/\min_i\{N_i\}$). We train for 200 epochs and use the top-1 accuracy as the evaluation metric. We compare FASA with two state-of-the-art feature-level augmentation approaches M2M~\cite{kim2020m2m} and Chu \etal~\cite{chu2020feature}.
% We do not compare with~\cite{Yin_2019_CVPR,liu2020deep} because they only report results on face recognition and person Re-ID, and no public codes are available for comparisons in other domains. 
When comparing with M2M~\cite{kim2020m2m}, we use LDAM~\cite{cao2019learning} loss as the baseline for a fair comparison. We use the same backbones of~\cite{kim2020m2m,chu2020feature} to experiment under the same settings.

From Table~\ref{tab:cifar100lt}, we observe that our FASA brings 1.7$\%$ accuracy improvement over the LDAM~\cite{cao2019learning} baseline and is comparable with M2M~\cite{kim2020m2m}. When applied to the stronger baseline De-confound-TDE~\cite{tang2020long}, FASA's benefits still hold. FASA is also better performing than Chu \etal~\cite{chu2020feature} with the cross entropy loss.
Other than that, our FASA is more time-efficient than M2M~\cite{kim2020m2m} and Chu \etal~\cite{chu2020feature} because they need extra time for classifier pre-training while FASA can be applied online.

%% file: sections/5_conclusion.tex
% !TEX root = ../main.tex
\section{Conclusion}

We have presented a simple yet effective method, Feature Augmentation and Sampling Adaptive (FASA), to address the class imbalance issue in the long-tailed instance segmentation task.
FASA generates virtual features on the fly to provide more positive samples for rare categories, and leverages a loss-guided adaptive sampling scheme to avoid over-fitting.
FASA achieves consistent gains on the Mask R-CNN baseline under different backbones, learning schedules, data samplers and loss functions, with minimal impact to the training efficiency. It shows large gains for rare classes without compromising the performance for common and frequent classes. Notably, as an orthogonal component, it improves other more recent methods such as RFS, EQL, cRT, BAGS and Seesaw.
Compelling results are shown on two challenging instance segmentation datasets, LVIS v1.0 and COCO-LT, as well as an imbalanced image classification benchmark, CIFAR-LT-100.

\noindent \textbf{Acknowledgements.}
This study is supported under the RIE2020 Industry Alignment Fund – Industry Collaboration Projects (IAF- ICP) Funding Initiative, as well as cash and in-kind contribution from the industry partner(s).

%% file: sections/6_appendix.tex
% !TEX root = ../main.tex

\clearpage

\begin{center}
    \section*{Appendix}
    % \vspace{-36pt}
\end{center}

\appendix

In the supplementary materials, we discuss the ablation studies and implementation details that are not elaborated in the main paper. Section \ref{sec:error_analysis} highlights our FASA largely reduces the classification error.
Section \ref{sec:ablation_sampling} analyzes some ablation studies of the adaptive feature sampling module.
% Section \ref{sec:visulaization_sampling} further provides the visualization result of adaptive feature sampling.
Section \ref{sec:speed} validates the memory- and time-efficiency of FASA.
Section \ref{sec:visulize_group} presents the clustering results used in the feature space.
Section \ref{sec:new_metric} provides the performance of FASA under the recently proposed \fixedap and \appool metrics~\cite{dave2021evaluating}.
Section \ref{sec:implementation} reports our implementation details of feature augmentation methods.
Section \ref{sec:comparison_nms_resampling} compares FASA with another instance-level re-sampling based approach NMS re-sampling~\cite{wu2020forest}.
Section \ref{sec:result_visualization} shows the visualization result of FASA.

\section{Error Analysis of Long-Tailed Instance Segmentation: Classification Error Dominates}
\label{sec:error_analysis}
\begin{table*}[t]
    \centering
    \tablestyle{12pt}{1.2}
	\caption{\small{Error analysis of Mask R-CNN~\cite{he2017mask} with and without FASA. We use the TIDE~\cite{daniel2020tide} toolbox and report the six error types (\%) on LVIS v1.0 \textit{validation} set: classification error ($\mathrm{E}_{\mathrm{cls}}$), location error ($\mathrm{E}_{\mathrm{loc}}$), both classification and location error ($\mathrm{E}_{\mathrm{both}}$), duplicate detection error ($\mathrm{E}_{\mathrm{dupe}}$), background error $\mathrm{E}_{\mathrm{bkg}}$ and missed ground truth error ($\mathrm{E}_{\mathrm{miss}}$). We observe the dominance of $\mathrm{E}_{\mathrm{cls}}$ for Mask R-CNN, and hence apply FASA only to the classification branch of Mask R-CNN at minimum cost. This leads to big improvements in $\mathrm{E}_{\mathrm{cls}}$ already. We expect more gains in $\mathrm{E}_{\mathrm{cls}}$ and other error metrics by augmenting other branches of Mask R-CNN if given more computational budget.}}
	\label{tab:ablation_error_analysis}
	\vspace{-6pt}
    \begin{tabular}{ccccccc}
    Method      & $\mathrm{E}_{\mathrm{cls}}$  & $\mathrm{E}_{\mathrm{loc}}$  & $\mathrm{E}_{\mathrm{both}}$ & $\mathrm{E}_{\mathrm{dupe}}$ & $\mathrm{E}_{\mathrm{bkg}}$  & $\mathrm{E}_{\mathrm{miss}}$ \\
    \midrule
    Mask R-CNN  & 25.10          & 6.71  & 0.59  & 0.36  & 3.17  & 7.15  \\
     + FASA     & 20.74          & 6.80  & 0.50  & 0.41  & 3.48  & 7.29  \\
    \midrule
    $\Delta$ & \textbf{-4.36} & +0.09 & -0.09 & +0.05 & +0.29 & +0.14 \\
    \end{tabular}
	\vspace{-6pt}
\end{table*}

In the main paper, we apply FASA only to the classification branch of Mask R-CNN. But why the choice? How does this simple mechanism impact performances of other branches like detection? To answer such questions, we need a detailed error analysis other than the default, single metric mean-average precision (mAP).
We use the recent TIDE~\cite{daniel2020tide} toolbox to report six error metrics for long-tailed instance segmentation.

Table~\ref{tab:ablation_error_analysis} outlines the six error metrics on large-scale LVIS dataset. We see that for the Mask R-CNN baseline~\cite{he2017mask}, \textbf{classification error} is the main bottleneck for long-tailed instance segmentation when compared to other error types,~\eg,~localization error. This explains our use of FASA to the classification branch of Mask R-CNN. Obviously, augmenting classification branch only will not incur a high cost. Performance-wise, we do see a significant reduction in classification error (from $25.10$\% to $20.74$\%) without deteriorating other errors much. With more budget, one could apply FASA to augment other branches of Mask R-CNN with the hope of more gains in all error metrics.

\section{Ablation on Adaptive Feature Sampling}
\label{sec:ablation_sampling}
\subsection{Initial Feature Sampling Probabilities}
To initialize our adaptive virtual feature sampling process, we assigned class-wise sampling probabilities based on inverse class frequency. This scheme favors rare-class feature augmentation at the beginning, and does not rely on too many assumptions about skewed data distribution. Another sampling scheme that has minimal assumption of data distribution is based on uniform class distribution.

Table~\ref{tab:ablation_init} compares the two schemes empirically. We see that both the uniform initialization and the inverse class frequency initialization boost the performance compared with the no augmentation baseline.
Overall, the inverse class frequency initialization approach achieves better overall mask mAP. The \apr and \apf results of uniform initialization are slightly worse than initialization based on inverse class frequency. So we use the initialization based on inverse class frequency by default for its effectiveness and simplicity.

\begin{table}[t]
    \centering
    \tablestyle{3pt}{1.2}
	\caption{\small{Comparing different initialization schemes of our virtual feature sampling probabilities.
    AP, \apr, \apc and \apf refer to the mask mAP metrics (\%) for overall, rare, common and frequent class groups.
    The symbol `FS' denotes to our adaptive Feature Sampling module.
    }}
    \label{tab:ablation_init}
	\vspace{-6pt}
    \begin{tabular}{lccccc}
    FS & Initial sampling probability & AP & \apr & \apc & \apf \\
    \midrule
    \gxmark & \gxmark  & 22.3 & 12.7 & 21.7 & 26.9 \\
    \cmark & Uniform distribution  & 23.2 & 15.1 & \bf 23.4 & 26.6 \\
    \cmark & Inverse class frequency & \textbf{23.7} & \textbf{17.8} & 22.9 & \textbf{27.2} \\
    \end{tabular}
    \vspace{-6pt}
\end{table}

\subsection{Performance Metric for Sampling Adaptation}
In the main paper, we propose a virtual feature sampling approach that is adapted to the validation loss rather than validation metric. This is to avoid the large computational cost from frequent metric evaluation on large-scale dataset. Concretely, evaluating the validation metric of mAP on the dataset takes nearly 45 minutes, which is very expensive if we were to conduct it in each epoch. To test how much we can gain from adapting to the true performance metric, we compare the use of the two supervisory signals on a smaller task. We choose the long-tailed image classification task on CIFAR-100-LT~\cite{cao2019learning}, a much smaller dataset than LVIS. Evaluating per-class accuracy is as efficient as evaluating the loss on CIFAR-100-LT validation set.
Table~\ref{tab:ablation_metric} shows a marginal improvement from using the performance metric, which when translated to large-scale dataset, may not be worth the large cost for metric evaluation.

\begin{table}[t]
    \centering
    \tablestyle{3pt}{1.2}
	\caption{\small{Comparing the use of different performance metrics (validation loss vs. metric) for adaptive feature sampling on CIFAR-100-LT~\cite{cao2019learning} dataset. The average and standard deviation of classification accuracy are from 3 runs.}
	}
	\label{tab:ablation_metric}
	\vspace{-6pt}
    \begin{tabular}{cc}
    Perf. measurement & Accuracy ($\%$) \\
    \midrule
    Validation loss     & 43.7 $\pm$ 0.25  \\
    Validation metric & \textbf{43.9 $\pm$ 0.38}  \\
    \end{tabular}
    \vspace{-6pt}
\end{table}

\subsection{Group-wise vs. Class-wise Adaptation}
By default, we adjust the feature sampling probability for each class group rather than each class. One of the reasons is that some classes may be missing for performance evaluation,~\eg,~on LVIS validation set. This makes it impossible for loss-adapted sampling probability adjustment for every class. But what if all classes are available for evaluation, appearing on both training and validation sets? 

We again test on the CIFAR-100-LT~\cite{cao2019learning} dataset that meets the requirement. In this case, class grouping is not a \textit{must} anymore, and our goal is to see if group-wise sampling adaptation still holds its benefits over class-wise sampling adaptation.
Table~\ref{tab:ablation_classwise} gives a positive answer.
We observe that group-wise sampling adaptation performs better and has a lower variance since it relies on the stabler group-wise loss average rather than the noisy per-class loss.

\begin{table}[t]
    \centering
    \tablestyle{3pt}{1.2}
	\caption{\small{Comparing the group-wise and class-wise feature sampling adaptation on CIFAR-100-LT~\cite{cao2019learning} dataset. The average and standard deviation of classification accuracy are from 3 runs.}
	}
	\label{tab:ablation_classwise}
	\vspace{-6pt}
    \begin{tabular}{cc}
    Sampling adaptation & Accuracy ($\%$) \\
    \midrule
    Group-wise & \textbf{43.7 $\pm$ 0.25}  \\
    Class-wise & 43.3 $\pm$ 0.85  \\
    \end{tabular}
    \vspace{-6pt}
\end{table}

\begin{table}[t]
    \centering
    \tablestyle{3pt}{1.2}
	\caption{\small{Comparison of training memory $\mathrm{M}_{\mathrm{train}}$ and training time $\mathrm{T}_{\mathrm{train}}$ required, with and without FASA on LVIS v1.0.}}
	\label{tab:time}
	\vspace{-6pt}
    \begin{tabular}{cccc}
    Method                          & FASA       & $\mathrm{M}_{\mathrm{train}}$ (GB) & $\mathrm{T}_{\mathrm{train}}$ (s/iter)\\
    \midrule
    \multirow{2}*{Mask R-CNN + RFS} & \gxmark     & 11.1             & 0.768 $\pm$ 0.04 \\
    ~                               & \cmark & 12.4             & 0.792 $\pm$ 0.05 \\
    \end{tabular}
	\vspace{-12pt}
\end{table}

\section{Speed Analysis}
\label{sec:speed}
Table~\ref{tab:time} further validates the memory- and time-efficiency of our FASA approach.
We see that FASA adds only a small amount of memory, which is used to maintain the online feature mean and variance of observed training samples. Thus the extra memory is constant and dependent only on the feature dimension. FASA is also found to incur a very small time cost.

\section{Visualizing Class Grouping Results}
\label{sec:visulize_group}
\begin{figure*}[!t]
\centering
\includegraphics[width=1.0\linewidth]{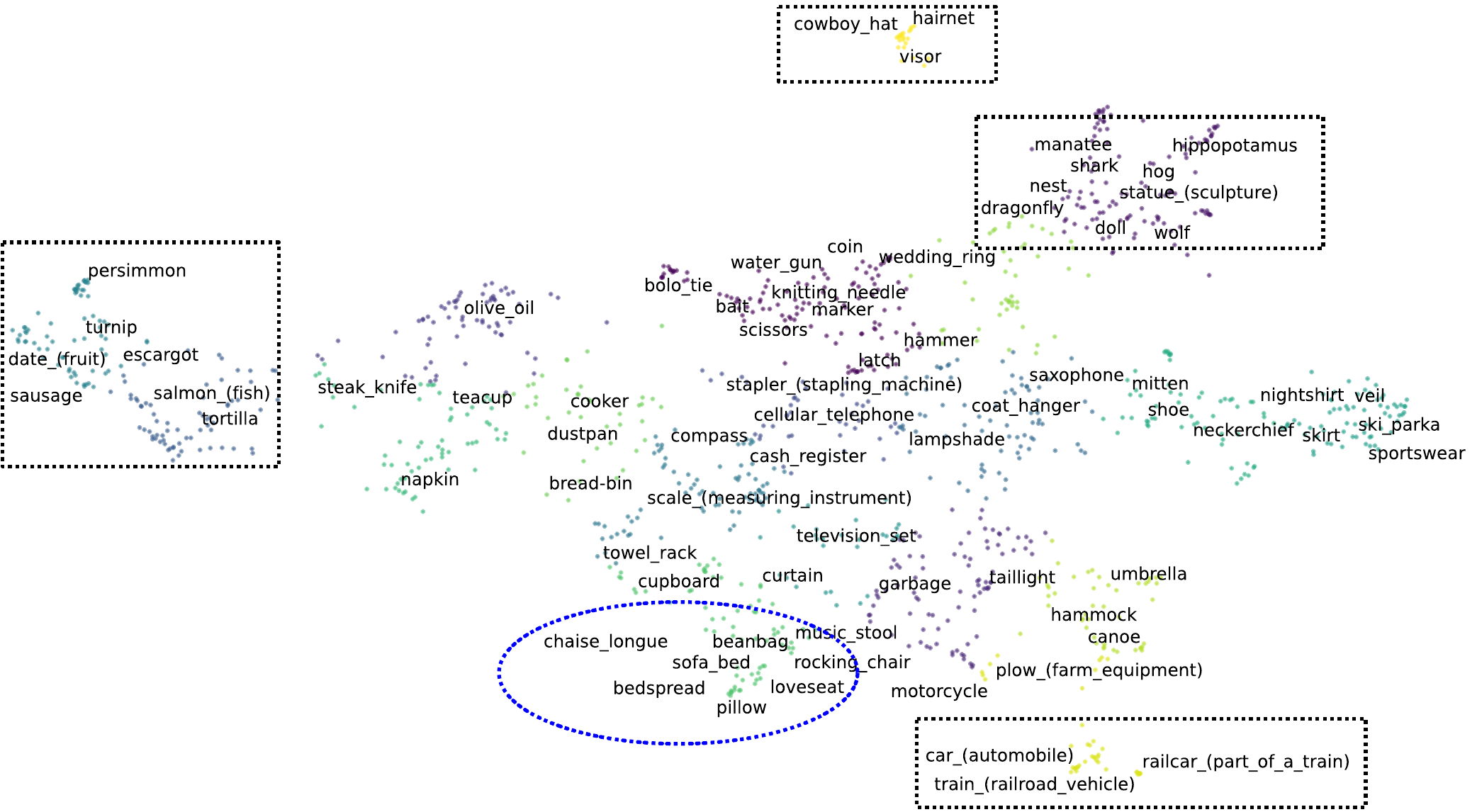}
\caption{t-SNE~\cite{maaten2008visualizing} visualization of class groups.
In the black dashed boxes, classes are often semantically related or visually similar.
In the blue ellipse, we find classes that exhibit strong co-occurrence,~\eg,~between `pillow' and `bedspread'.
}
\label{fig:clustering}
\end{figure*}

Recall our feature sampling probabilities are adjusted in a group-wise manner. The class groups are formed by the Mean-shift~\cite{cheng1995mean} clustering algorithm.
Figure~\ref{fig:clustering} presents some clustering results, where visually similar or semantically related classes stay close in the feature space (\eg,~“hairnet” and “visor”).
In addition, the co-occurrent classes also tend to stay close (\eg,~“pillow” and “loveseat”).
Intuitively, related classes are better suited to have their sampling probabilities adjusted together.

\section{Evaluation with \fixedap and \appool}
\label{sec:new_metric}
The mean average precision metric (denotes as \oldap in the following) is the default evaluation metric for the instance segmentation task~\cite{lin2014microsoft,gupta2019lvis}.
Recently, Dave~\etal~\cite{dave2021evaluating} argued that the \oldap metric is sensitive to changes in cross-category ranking, and introduced two complementary metrics \fixedap and \appool to replace \oldap for LVIS~\cite{gupta2019lvis} dataset.
Dave~\etal~\cite{dave2021evaluating} found that some methods improve \oldap but have less impact on \fixedap and \appool.
The \oldap metric limits the maximum detection results per image, resulting in cross-category competition.
To address this issue, the \fixedap metric limits the maximum detection results per class on the dataset instead.
To highlight the score calibration property, Dave~\etal~\cite{dave2021evaluating} also proposed \appool metric that is class-agnostic and evaluates detection results across all categories together.
Since \appool is class-agnostic, the evaluation is influenced more heavily by frequent classes rather than rare classes.

We provide the experimental results of FASA under the \fixedap and \appool in Table~\ref{tab:new_metric}.
We observe that FASA consistently boosts the performance under \fixedap and \appool, especially for the rare categories.
For \fixedap, FASA improves overall AP and rare-class \apr by 1.1$\%$ / 1.7$\%$ respectively for Mask R-CNN and 1.3$\%$ / 2.4$\%$ for Cascade Mask R-CNN.
For \appool, FASA still obtains 2.9$\%$ gains in \apr for Mask R-CNN and 2.0$\%$ for  Cascade Mask R-CNN. Since \appool is mainly affected by the frequent class, the overall AP improvements are small.
The \fixedap and \appool results also demonstrate that FASA largely improves the performance of rare classes without compromising the common classes and frequent classes.

\begin{table*}[t]
    \centering
    \tablestyle{8pt}{1.1}
	\caption{\small{Results of Cascade Mask R-CNN~\cite{cai2019cascade} with and without FASA under the recently proposed \fixedap and \appool metric~\cite{dave2021evaluating}. We report the results on the LVIS~\cite{gupta2019lvis} \textit{validation} set.
	AP, \apr, \apc and \apf refer to the mask mAP metrics (\%) for overall, rare, common and frequent class groups.
	The symbol \oldap refers to the standard mean average precision (mAP) metric.
	We observe FASA offers consistent performance boost under the \fixedap and \appool, especially for the rare categories.
	All the models use the ResNet-101~\cite{he2016deep} backbone and Repeat Factor Sampling~(RFS)~\cite{gupta2019lvis}.
	}}
	\label{tab:new_metric}
	% \vspace{-6pt}
	\if 0
    \begin{tabular}{
    cc
    x{20}x{20}x{6}x{6}
    x{20}x{20}x{6}x{6}
    x{20}x{20}x{6}x{6}
    }
    &
    & \multicolumn{4}{c}{\oldap}
    & \multicolumn{4}{c}{\fixedap}
    & \multicolumn{4}{c}{\appool}
    \\
    \cmidrule(lr){3-6}
    \cmidrule(lr){7-10}
    \cmidrule(lr){11-14}
    Method     & FASA   & AP   & \apr & \apc & \apf & AP   & \apr & \apc & \apf & AP   & \apr & \apc & \apf \\
    \midrule
    \multirow{2}*{Mask R-CNN~\cite{he2017mask}}  & \gxmark & 
    24.4        & 13.2 & 24.7 & 30.3 & 27.1 & 20.3 & 26.9 & 30.3 & 27.2 & 9.0 & 22.5 & 27.5 \\
    ~  & \cmark & 
    \bf 26.3 \td{+1.9} & \bf 19.1 \td{+5.9} & \bf 25.4 & \bf 30.6 & \bf 28.2 \td{+1.1} & \bf 22.0 \td{+1.7} & \bf 28.3 & \bf 30.9 & \bf 27.4 \td{+0.2} & \bf 11.9 \td{+2.9} & \bf 23.0 & \bf 27.8 \\
    \midrule
    \multirow{2}*{Cascade Mask R-CNN~\cite{cai2019cascade}}  & \gxmark & 
    25.4        & 13.7 & 24.8 & 31.4 & 28.7 & 22.2 & 28.3 & 32.0 & 28.9 & 10.4 & 24.2 & 29.4 \\
    ~  & \cmark & 
    \bf 27.7 \td{+2.3} & \bf 19.8 \td{+5.9} & \bf 27.3 & \bf 31.6 & \bf 30.0 \td{+1.3} & \bf 24.6 \td{+2.4} & \bf 29.8 & \bf 32.4 & \bf 29.2 \td{+0.3} & \bf 12.4 \td{+2.0} & \bf 25.0 & \bf 29.6 \\
    \end{tabular}
    \fi
    \begin{tabular}{
    cc
    x{20}x{20}x{6}x{6}
    x{20}x{20}x{6}x{6}
    }
    &
    & \multicolumn{4}{c}{\fixedap}
    & \multicolumn{4}{c}{\appool}
    \\
    \cmidrule(lr){3-6}
    \cmidrule(lr){7-10}
    Method     & FASA   & AP   & \apr & \apc & \apf & AP   & \apr & \apc & \apf \\
    \midrule
    \multirow{2}*{Mask R-CNN~\cite{he2017mask}}  & \gxmark & 27.1 & 20.3 & 26.9 & 30.3 & 27.2 & 9.0 & 22.5 & 27.5 \\
    ~  & \cmark & \bf 28.2 \td{+1.1} & \bf 22.0 \td{+1.7} & \bf 28.3 & \bf 30.9 & \bf 27.4 \td{+0.2} & \bf 11.9 \td{+2.9} & \bf 23.0 & \bf 27.8 \\
    \midrule
    \multirow{2}*{Cascade Mask R-CNN~\cite{cai2019cascade}}  & \gxmark & 28.7 & 22.2 & 28.3 & 32.0 & 28.9 & 10.4 & 24.2 & 29.4 \\
    ~  & \cmark & \bf 30.0 \td{+1.3} & \bf 24.6 \td{+2.4} & \bf 29.8 & \bf 32.4 & \bf 29.2 \td{+0.3} & \bf 12.4 \td{+2.0} & \bf 25.0 & \bf 29.6 \\
    \end{tabular}
\end{table*}

\section{Implementation Details}
\label{sec:implementation}

Our implementation is based on the Mask R-CNN~\cite{he2017mask} backbone, which is the ImageNet pre-trained ResNet-50~\cite{he2016deep} with a FPN~\cite{lin2017feature} neck and a box head with two sibling fully connected layers for RoI classification and regression.
We apply random horizontal image flipping and multi-scale jittering with the smaller image sizes (640, 672, 704, 736, 768, 800) in all experiments.
All models are trained with standard SGD on 8 NVIDIA V100 GPUs.
We follow the default settings in MMDetection~\cite{mmdetection} to set other hyper-parameters such as learning rates and training schedules.

Here we also describe the implementation details of the feature augmentation methods listed in Table~2 of the main paper.
We detail the hyper-parameters of these approaches and our searched optimal choices in Table~\ref{tab:supp_tuning}.

\begin{table*}[t]
\centering
\caption{Parameters tuned for the feature augmentation methods in Table~2 of the main paper.}
\label{tab:supp_tuning}
\tablestyle{6pt}{1.1}
\begin{tabular}{lc@{\hskip 5pt} lc}
Method & Param & Description & Value \\
\midrule
\multirow{2}*{MoEx, CVPR'21~\cite{li2020feature}} & $p$ & MoEx probability & 1.0 \\
~ & $\epsilon$ & Epsilon constant for standard deviation & $1e^{-5}$ \\
\midrule
\multirow{2}*{Liu~\etal, CVPR'20~\cite{liu2020deep}} & $s$ & Scaling factor & 20 \\
~ & $m_{a}$ & Angular margin & 0.1 \\
\midrule
\multirow{2}*{Chu~\etal, ECCV'20~\cite{chu2020feature}} & $\mathcal{T}_{s}$ & Threshold to extract the class-specific features & 0.3 \\
~ & $\mathcal{T}_{g}$ & Threshold to extract the class-generic features & 0.6 \\
\midrule
\multirow{2}*{Yin~\etal, CVPR'19~\cite{Yin_2019_CVPR}} & $\alpha_{\text{recon}}$ & Coefficient of the reconstruction loss & 0.5 \\
~ & $\alpha_{\text{reg}}$ & Coefficient of the regularization loss & 0.25
\end{tabular}
\vspace{-12pt}
\end{table*}

\smallsec{SMOTE~\cite{chawla2002smote}.}
The SMOTE algorithm interpolates neighboring features (\ie,~feature embeddings of region proposals) in the feature space.
Specifically, for given features $x_{i}$, we consider the $k=5$ nearest neighbours $\{x_{j}\}$ based on cosine feature distance. Then we interpolate new features as:
\begin{equation}
\begin{aligned}
    \hat{x} = \lambda \cdot x_{i} + (1 - \lambda) \cdot x_{j}, \\
\end{aligned}
\end{equation}
where $\lambda$ is a random value in $(0, 1]$.

\smallsec{MoEx~\cite{li2020feature}.}
Similar to SMOTE~\cite{chawla2002smote}, the MoEx is also an interpolation-based augmentation method. 
As MoEx was developed for the image classification task, we transfer it to the instance segmentation task with the following modification:
1) we applied MoEx augmentation in the classifier of the Mask R-CNN~\cite{he2017mask} framework,
2) we searched the optimal value of parameters on the LVIS dataset and the results are shown in Table~\ref{tab:supp_tuning}.

\smallsec{InstaBoost~\cite{fang2019instaboost}.}
InstaBoost is an adaptive copy-and-paste FA method based on a location probability map. Since InstaBoost was already developed for the instance segmentation task, we use its default hyper-parameters.

\if 0
\smallsec{Ghiasi~\etal~\cite{ghiasi2020simple}.}
Ghiasi~\etal~\cite{ghiasi2020simple} is the state-of-the-art copy-and-paste augmentation approach for the instance segmentation task.
The experiment results of Ghiasi~\etal~\cite{ghiasi2020simple} were based on the EfficientDet~\cite{tan2020efficientdet} framework.
For a fair comparison with other methods, we re-implemented it on the Mask R-CNN~\cite{he2017mask} using the default hyper-parameters.
\fi

\smallsec{Liu~\etal~\cite{liu2020deep}.}
Liu~\etal~\cite{liu2020deep} propose to transfer the angular distribution of face recognition loss such as CosFace~\cite{wang2018cosface} or ArcFace~\cite{deng2019arcface}.
We select ArcFace for re-implementation:
\begin{equation}
\label{eq:arc_face}
\begin{aligned}
L=-\frac{1}{N} \sum_{n=1}^{N} \log \frac{e^{s\left(\cos \left(\theta_{y}+\alpha_{y}+m_{a}\right)\right)}}{e^{s\left(\cos \left(\theta_{y}+\alpha_{y}+m_{a}\right)\right)}+\sum_{j \neq y}^{C} e^{s\left(\cos \left(\theta_{j}+\alpha_{y}\right)\right)}}
\end{aligned}    
\end{equation}
The symbol $\theta_{y}$ refers to the angle between the input feature and the weight of the classifier.
The symbol $\alpha_{y}$ means the extra angular that transfers from head class to tail class.
As shown in Eq~\eqref{eq:arc_face}, two parameters are involved: the symbol $s$ means the scaling factor applied to logit, and the symbol $m_{a}$ refers to the angular margin.
We tune these two parameters and show the results in Table~\ref{tab:supp_tuning}.

% We observed that when applied ArcFace~\cite{deng2019arcface} on instance segmentation task, the normalization of input features and the weight of classifier can slightly boost the performance. The similar observations from ~\cite{dave2021evaluating,wang2020seesaw} confirm the help of normalized classifier head. 
We observed that since the instance segmentation task has to deal with the special \textit{background} class, the margin-based ArcFace loss is unfortunately very sensitive to hyper-parameter choices of $m_{a}$.
So margin-based augmentation~\cite{liu2020deep} does not perform well on LVIS.
Different from Liu~\etal~\cite{liu2020deep}, our FASA is not limited to the form of loss functions.

\smallsec{Chu~\etal~\cite{chu2020feature}.}
Chu~\etal~\cite{chu2020feature} mixed the class-specific features of each class and the corresponding class-generic features of `confusing' classes to synthesize new data samples. The definitions of `class-generic' and `class-specific' are based on the threshold masking of class activation map (CAM)~\cite{zhou2016learning}. For each real sample in the tail class, the authors sample $N_{a}$ images from its $N_{f}$ confusing classes.

During the re-implementation, we found that the difference in batch size between the classification and instance segmentation tasks limited the performance of Chu~\etal\cite{chu2020feature} when transferred to the instance segmentation task.
The classification task has a large batch size (\eg, 128) that can meet the demand of picking confusing categories (\eg, $N_{a}=N_{f}=3$). However, instance segmentation models are limited by small batch size (\eg, 2) and there is no guarantee that the top confusing categories will appear in the same batch.
Compared with Chu~\etal\cite{chu2020feature}, our FASA builds feature banks for each category to cache the features of the previous batch, thus getting rid of the small batch limitation.

\smallsec{Yin~\etal~\cite{Yin_2019_CVPR}.}
Yin~\etal~\cite{Yin_2019_CVPR} is a feature augmentation method designed for the face recognition task.
A total of three loss functions are included: face classification loss $\mathcal{L}_{sfmx}$, reconstruction loss $\mathcal{L}_{\text{recon}}$ and regularization loss $\mathcal{L}_{\text{reg}}$.
The reconstruction loss $\mathcal{L}_{\text{recon}}$ is critical to train the discriminative feature encoder and decoder.
To transfer into the instance segmentation task, we apply the reconstruction loss to the feature embedding of each positive region proposal.
Besides, Yin~\etal~\cite{Yin_2019_CVPR} need a two-stage training pipeline.
In the first stage, the authors fix the backbone and generate new feature samples to train the classifier.
In the second stage, the authors fix the classifier and update the other components.
Such a two-stage approach introduces additional training time cost.
Compared to Yin~\etal~\cite{Yin_2019_CVPR}, our FASA leverages end-to-end training and therefore more efficient.

\begin{figure*}[ht]
\centering
\includegraphics[width=\linewidth]{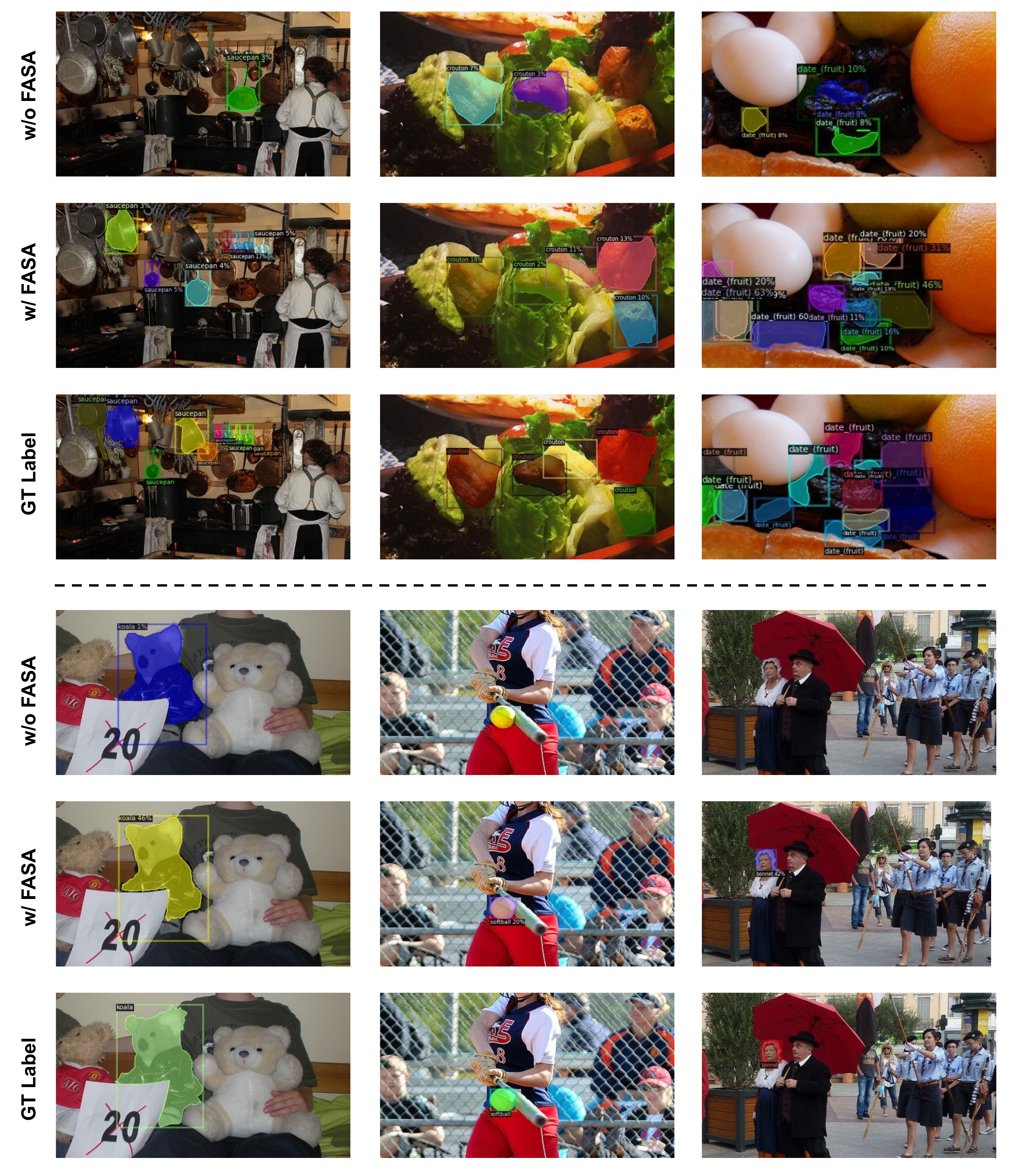}
\vspace{-12pt}
\caption{Prediction results of Mask R-CNN framework without and with FASA on the LVIS v1.0 \textit{validation} set. We select six rare classes `saucepan', `crouton', `date (fruit)', `koala', `softball' and `bonnet' to visualize. We observe that with the help of FASA, Mask R-CNN exhibits more correct classification results than the baseline.
}
\label{fig:vis}
\vspace{-12pt}
\end{figure*}

\section{Comparison with NMS Re-sampling~\cite{wu2020forest}}
\label{sec:comparison_nms_resampling}
NMS Re-sampling~\cite{wu2020forest} is proposed to adjusting the NMS threshold for different categories during the training. Specifically, the NMS thresholds for the frequent/common/rare categories are set as \{0.7, 0.8, 0.9\} with the increasing trend. Such a mechanism is beneficial to preserve more region proposals from the rare classes and suppress the number of proposals from frequent classes.

\begin{table}[t]
    \centering
    \tablestyle{3pt}{1.2}
	\caption{\small{Comparing our FASA with NMS Re-sampling~\cite{wu2020forest} on LVIS v1.0 \textit{validation} set.
	The symbol `NR' denotes the NMS Resampling approach.
    AP, \apr, \apc and \apf refer to the mask mAP metrics (\%) for overall, rare, common and frequent class groups.
    }}
    \label{tab:compare_nms}
	\vspace{-6pt}
    \begin{tabular}{lccccc}
    NR & FASA & AP & \apr & \apc & \apf \\
    \midrule
    \gxmark & \gxmark & 19.3 & 1.2 & 17.4 & \bf 29.3 \\
    \gxmark & \cmark  & 22.6 & 10.2 & 21.6 & 29.2 \\
    \cmark  & \gxmark & 21.7 & 8.6 & 20.4 & 29.0 \\
    \cmark  & \cmark  & \bf 22.9 & \bf 11.1 & \bf 21.8 & 29.2 \\
    \end{tabular}
    \vspace{-6pt}
\end{table}

We compare the FASA with NMS Re-sampling in Table~\ref{tab:compare_nms}.
The first line refers to the Mask R-CNN~\cite{he2017mask} baseline without any re-sampling or augmentation method.
From the second and the third line, we see that both FASA and NMS Re-sampling achieve better performance than the baseline method.
FASA performs slightly better than NMS Re-sampling, especially for the rare classes.
We believe such a performance gap is due to NMS Re-Sampling is mainly in adjusting the sampling weights of the current data samples, while FASA can further generate \textit{new} virtual samplers.
Also, the bottom results demonstrate that FASA as an orthogonal module can combine with NMS Re-sampling to further boost the performance.

\section{Result Visualization}
\label{sec:result_visualization}

To better interpret the result, we show the segmentation results of the selected rare classes in Figure~\ref{fig:vis}.
% From the top row to the third row, we show the prediction results of Mask R-CNN without/with FASA and the ground truth label.
We observe that without FASA, the prediction scores for rare classes are small or even missed.
On the contrary, with the help of our FASA, the classification results of the rare classes become accurate.